%% file: aaai22.tex
\def\year{2022}\relax
\documentclass[letterpaper]{article}
\usepackage{aaai22} 
\usepackage{times} 
\usepackage{helvet} 
\usepackage{courier} 
\usepackage[hyphens]{url} 
\usepackage{graphicx} 
\usepackage{graphicx}  
\usepackage{natbib}  
\usepackage{caption}  
\DeclareCaptionStyle{ruled}{labelfont=normalfont,labelsep=colon,strut=off}
\usepackage[linesnumbered,noend]{algorithm2e}
\usepackage{xcolor}
\usepackage{booktabs}       
\usepackage{amsfonts}       
\usepackage{nicefrac}       
\usepackage{microtype}      
\usepackage{xcolor}         

\graphicspath{{figures/}}
\usepackage{mathtools}
\usepackage{booktabs}

\usepackage{subcaption}
\usepackage[font={small,it},skip=1pt]{caption}
\usepackage{tabularx}
\usepackage{color}
\usepackage{graphics}
\usepackage{enumitem}
\usepackage{multirow}

\usepackage{cleveref}
\newcommand{\MDP}{\textsf{MDP}}
\newcommand{\SCM}{\textsf{SCM}}

\definecolor{amber}{rgb}{1.0, 0.75, 0.0}
\newcommand{\var}[1]{\textsf{#1}}

\newcommand{\functionname}[1]{\texttt{#1}}

\newcommand{\x}[1]{\textbf{#1}}
\newcommand{\RQ}[1]{\textbf{RQ#1}}

\definecolor{gray}{rgb}{0.7,0.7,0.7}
\usepackage{pifont}
\newcommand{\cmark}{\ding{51}}%
\newcommand{\xmark}{\textcolor{gray}{\ding{55}}}%

\newcommand{\xxmark}{\ding{56}\xspace}
\newcommand{\cxmark}{\ding{52}\xspace}

\newcommand{\toolname}{\textsc{FastAR}\xspace}

\usepackage[compact]{titlesec}
\titlespacing{\section}{0pt}{0ex}{0ex}
\titlespacing{\subsection}{0pt}{0ex}{0ex}
\usepackage{amsthm}
\newtheorem{Ques}{Ques}
\newcommand{\Qref}[1]{\textbf{Ques~\ref{#1}}}

\newtheorem*{Justify}{Ans}

\title{Amortized Generation of Sequential \\ Algorithmic Recourses for Black-box Models}

\author{%
  Sahil Verma,\textsuperscript{\rm 1,2}
  Keegan Hines,\textsuperscript{\rm 1}
  John P. Dickerson\textsuperscript{\rm 1,3}
}
\affiliations{
    \textsuperscript{\rm 1} Arthur AI \\ 
    \textsuperscript{\rm 2} University of Washington \\ 
    \textsuperscript{\rm 3} University of Maryland \\
    vsahil@cs.washington.edu, keegan@arthur.ai, john@arthur.ai
    }

\begin{document}

\maketitle

\begin{abstract}
  \input{abstract}
\end{abstract}

\input{introduction}
\input{desiderata}
\input{example_shorter}

\input{algorithm}     
\input{evaluation}
\input{conclusion}


\small
\bibliography{refs.bib}



\clearpage
\appendix
\normalfont
\input{appendix}

\end{document}

%% file: abstract.tex
Explainable machine learning (ML) has gained traction in recent years due to the increasing adoption of ML-based systems in many sectors.  
\emph{Algorithmic Recourses} (ARs) provide ``what if'' feedback of the form ``if an input datapoint were $x'$ instead of $x$, 
then an ML-based system's output would be $y'$ instead of $y$.''  
ARs are attractive due to their actionable feedback, amenability to existing legal frameworks, 
and fidelity to the underlying ML model.  
Yet, current AR approaches are single shot---that is, they assume $x$ can change to $x'$ in a single time period.  
We propose a novel stochastic-control-based approach that generates \emph{sequential} ARs, 
that is, ARs that allow $x$ to move stochastically and sequentially across intermediate states to a final state $x'$.  
Our approach is model agnostic and black box. 
Furthermore, the calculation of ARs is amortized such that once trained, it applies to multiple datapoints without the need for re-optimization.  
In addition to these primary characteristics, our approach admits optional desiderata 
such as adherence to the data manifold, respect for causal relations, and sparsity---identified by past research as desirable properties of ARs.  
We evaluate our approach using three real-world datasets and show successful generation 
of sequential ARs that respect other recourse desiderata. 

%% file: introduction.tex
\section{Introduction}
\label{sec:intro}

Machine learning (ML) models are increasingly used to make predictions in systems that directly or indirectly impact humans. This includes critical applications like healthcare~\citep{medical-treatment-ml}, finance~\citep{credit-risk-ml}, hiring~\citep{hiring-ml}, and parole~\citep{parole-ml}. 
To understand ML models better and to promote their equitable impact in society, it is necessary to assess stakeholders'---both expert~\citep{Holstein19:Improving} and layperson~\citep{Saha20:Measuring}---comprehension of and needs for general observability into their systems~\citep{poursabzi2021manipulating,ehsan2021expanding}.
The nascent Fairness, Accountability, Transparency, and Ethics in machine learning (aka ``FATE ML'') community conducts research to develop methods to detect (and counteract) bias in ML models, develop techniques that make complex models explainable, and propose policies to advise and adhere to the regulations of algorithmic decision-making (see \Cref{sec:background}). 
Here, we focus on ML model explainability. 

Research in explainable ML is bifurcated. One high-level approach aims to develop inherently interpretable models such as decision trees and linear models~\citep{Rudin19:Stop}. 
Another high-level approach aims to utilize existing complex classification techniques (such as deep neural networks) but to bolster them with surrogate models that can render their predictions and/or internal processes understandable~\citep{xai-survey2}. 
This is achieved through explaining models holistically (global explanation) or single predictions from the model (local explanation). 

\textbf{Algorithmic Recourses (ARs).}  
ARs find the minimal change in a datapoint such that the ML model ends up classifying the new datapoint in the desired class. 
Such new datapoint(s) is termed as a counterfactual. 
(We provide an in-depth discussion of terminology in \Cref{sec:counter_vs_contra}.) 
For example, if an individual were denied a loan request, a recourse might tell them that their request would be approved if they were to increase their income by \$2000. 
ARs provide a precise recommendation and are therefore more actionable than other forms of local explainability like feature importance. 
Recent research in this area has aimed to ensure ARs are actionable and useful by incorporating additional desiderata into the recourse generation problem. 
As described in~\Cref{sec:desiderata}, these include notions of sparsity, causality, and realism of ARs, among others. 
What is needed~\citep[see, e.g.,][]{verma2020CFsurvey,Chou21:Counterfactuals,karimi2020survey} 
is a generalized approach that can accommodate such varied constraints and 
can also be computed efficiently.

\textbf{Operationalizing ARs.}  
We propose a novel approach (\toolname) for generating ARs by translating a given recourse generation problem into a Markov Decision Process (\MDP{}). 
\toolname aims to learn a policy that can generate ARs for given data distribution. 
Upon learning that policy once, it can generate ARs for multiple datapoints (from that distribution) without the need to re-optimize (which is required by most previous approaches; see~\Cref{sec:related}). 
Thus, \toolname \emph{amortizes} the cost of repeatedly computing ARs. 
\toolname also allows enforcing desirable properties of ARs, such as closeness to the training data distribution (data manifold), respect of causal relations between the features, and mutability and actionability of different features. 
\toolname works for \emph{black-box} models and is therefore \emph{model agnostic}. 

Via the learned policy, \toolname outputs ARs as a sequence of steps that lead an individual to a counterfactual state. 
To our knowledge, we are the first to leverage techniques from stochastic control to provide such \emph{sequential ARs}~\citep{Ramakrishnan_Lee_Albarghouthi_2020,naumann2021sequential}. 
That sequence can also adhere to particular \emph{sparsity} constraints (e.g., only one feature changing per step). 


Sequential and ``rolled out'' ARs have several advantages, directly addressing gaps 
identified by recent survey papers~\citep{verma2020CFsurvey,Chou21:Counterfactuals,karimi2020survey} 
and workshops~\citep{Ehsan21:Operationalizing}: 
1) action sparsity allows an individual to focus their effort on changing 
a small number of features at a time; and
2) presentation of ARs as a set of discrete and sequential steps is closer to real-world actions, rather than one-step continuous change, which most previous approaches do. 
\citet{singh_directive_2021} recently conducted a user-study with 54 participants, 
wherein each of them was presented with 15 scenarios and asked if they preferred one-shot or directed sequential AR in that scenario. 
When overall results were pulled, the study concluded a preference for sequential ARs with high confidence. 

\begin{figure}
    \centering
    \includegraphics[width=\columnwidth]{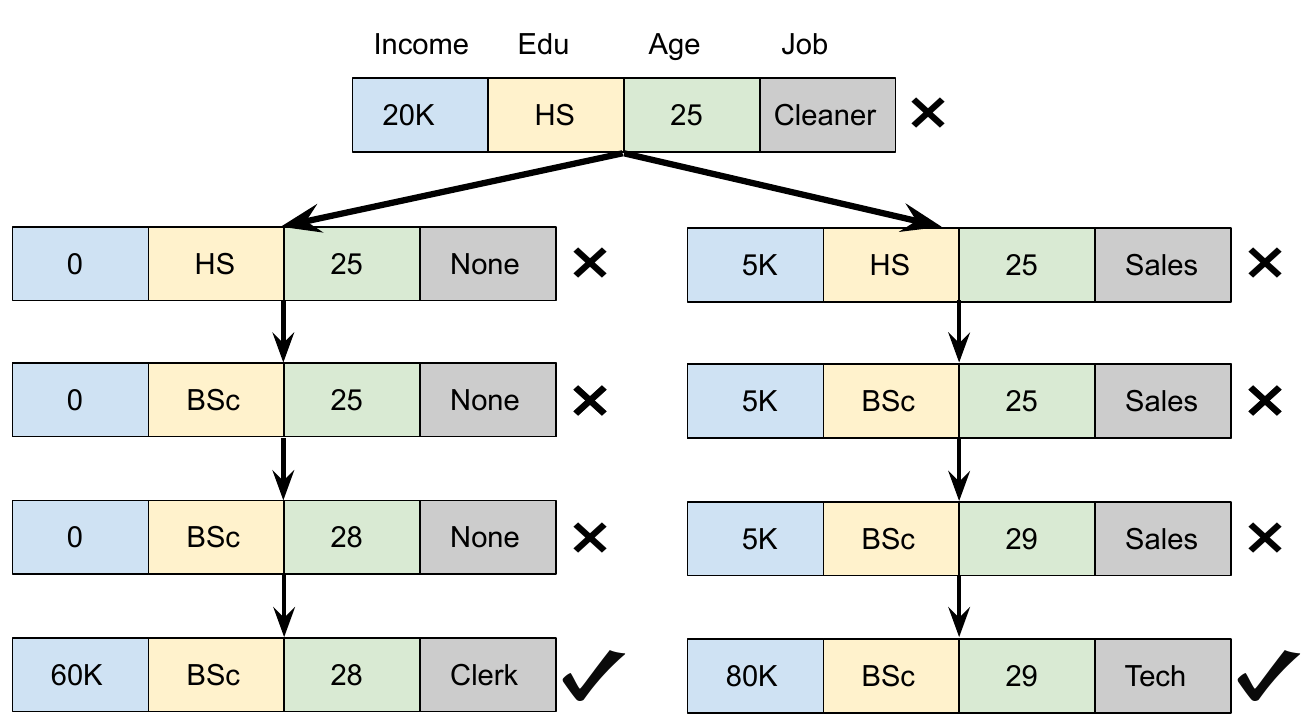}
    \caption{Example of Stochastic Algorithmic Recourses. 
    Starting with a datapoint (\xxmark denotes undesired class prediction), 
    \toolname can stochastically generate ARs that lead to different counterfactual states 
    (\cxmark denotes desired class prediction). }
    \label{fig:seq_ar}
\end{figure}

\Cref{fig:seq_ar} shows an example of sequential ARs which are generated for an individual whose loan request was denied (shown by \xxmark). 
Instead of a one-shot solution, \toolname delineates all intermediate steps to reach a counterfactual state (shown by \cxmark). 
\toolname also models the stochastic factors like the duration to complete a BSc degree, no or part-time job during the course, and the salary variance in the new job after graduation, 
which lead to different recourse paths and hence different counterfactual states (as shown in~\Cref{fig:seq_ar}). 

In summary, our contributions are:
\begin{enumerate}[noitemsep,topsep=0pt,parsep=0pt,partopsep=0pt,leftmargin=*]
 \item A novel algorithm that translates an AR problem into a Markov decision process (\MDP{}). 
 To the best of our knowledge, our stochastic-control-based approach is the first to address several roadblocks to using ARs in practice that have been identified by the 
 community~\citep{verma2020CFsurvey,Chou21:Counterfactuals,karimi2020survey}.
 \item The first approach that generates sequential and amortized ARs, and also works for black-box models. 
 \item An extensive evaluation with three real-world datasets and nine baselines. 
\end{enumerate}






%% file: desiderata.tex
\section{Desiderata of \emph{Practical} ARs}
\label{sec:desiderata}

The overarching goal of an AR is to provide practical guidance to an individual seeking to change their treatment (e.g., class label) by a deployed ML model. 
Apart from the necessary property of a AR having a desired class label, other desiderata have been identified in the literature, enumerated here:
\begin{itemize}[noitemsep,topsep=0pt,parsep=0pt,partopsep=0pt,leftmargin=*]
 \item \emph{Actionability}: ARs should only recommend changes to the features that are actionable~\citep{Ustun19:Actionable,Kanamori2020:DACE,dandl_multi-objective_2020}. 
 Actionable features are dataset and preference dependent. 
 
 \item \emph{Sparsity}: Social studies have argued that smaller explanations are more comprehensible to humans~\citep{Miller-xai:2019}. 
 Therefore ARs should make changes to a small set of features~\citep{van_looveren_interpretable_2020,karimi_model-agnostic_2020}.
 
 \item \emph{Data manifold}: To obey the correlations between features, their input domain, and to be realistic, ARs should adhere to the training data manifold~\citep{dhurandhar_model_2019,Kanamori2020:DACE,dandl_multi-objective_2020}.
 
 \item \emph{Causal constraints}: In order to adhere to real-world constraints in ARs, causal constraints between features must be respected. 
 They can encode facts like age cannot decrease or increase in education level increases age~\citep{mahajan_preserving_2020}. 
 
 
 \item \emph{Model-agnostic}: For wide-spread applicability, AR generating approaches should be model-agnostic~\citep{medina_comparison-based_2018,guidotti_local_2018}.
 
 \item \emph{Black-box models}: For applicability to proprietary ML models, AR generating approaches should work for black-box models~\citep{sharma_certifai_2019}.

 \item \emph{Amortized}: An \emph{amortized} approach can generate ARs for several 
 datapoints without optimizing separately for each of them. 
 Such an approach is effective for deployment~\citep{mahajan_preserving_2020}. 

    %
\end{itemize}

\toolname satisfies all the above desiderata. To the best of our knowledge, it is the 
first approach to do so (see~\Cref{tab:main-table}). 
The choice of action space helps produce ARs that consider actionability among features and are sparse. 
It only modifies the actionable features. 
Its ARs are realistic as they adhere to the training data manifold and respect causal relations between features. 
\toolname works for black-box models and, therefore, is model-agnostic. 
It learns a policy that can produce ARs for several input datapoints without the need to optimize again; and, therefore, generates amortized ARs. 


\begin{table*}
\caption{Desiderata comparison of various AR generating approaches. 
\toolname is the \x{first and only one} which satisfies all desiderata. }
 \label{tab:main-table}
 \centering
\resizebox{\textwidth}{!}{%
\begin{tabular}{ccccccccc}
 \toprule
Approach & Actionability & Sparsity & Agnostic & Black-box & Amortized & Manifold & Constraints \\
 \midrule
CFE Expl.~\citep{wachter_counterfactual_2017} & \xmark & \cmark & \xmark & \xmark & \xmark & \xmark & \xmark   \\
Recourse~\citep{Ustun19:Actionable} & \cmark & \cmark & \xmark & \xmark & \xmark  & \xmark & \xmark  \\ 
CEM~\citep{dhurandhar_model_2019} & \xmark & \cmark & \xmark & \xmark & \xmark  & \cmark & \xmark   \\ 
MACE~\citep{karimi_model-agnostic_2020} & \cmark & \cmark & \xmark & \xmark & \xmark & \xmark & \xmark  \\ 
DACE~\citep{Kanamori2020:DACE} & \cmark & \xmark & \xmark & \xmark & \xmark & \cmark & \xmark \\ 
DiCE~\citep{mothilal_explaining_2020}  & \cmark & \cmark & \xmark & \xmark & \xmark  & \xmark & \xmark\\ 
DiCE VAE~\citep{mahajan_preserving_2020} & \cmark & \xmark & \xmark & \xmark & \cmark  & \cmark & \cmark  \\ 
Spheres~\citep{medina_comparison-based_2018} & \xmark & \cmark & \cmark & \cmark & \xmark & \xmark & \xmark  \\ 
LORE~\citep{guidotti_local_2018} & \xmark & \cmark & \cmark & \cmark & \xmark & \xmark & \xmark  \\ 
Weighted~\citep{grath_interpretable_2018} & \xmark & \xmark & \cmark & \cmark & \xmark  & \xmark & \xmark  \\ 
CERTIFAI~\citep{sharma_certifai_2019} & \cmark & \xmark & \cmark & \cmark & \xmark  & \xmark & \xmark  \\ 
Prototypes~\citep{van_looveren_interpretable_2020} & \xmark & \cmark & \cmark & \cmark & \xmark  & \cmark & \xmark  \\ 
MOC~\citep{dandl_multi-objective_2020} & \cmark & \cmark & \cmark & \cmark & \xmark  & \cmark & \xmark  \\ 
\textbf{\toolname} & \cmark & \cmark & \cmark & \cmark & \cmark & \cmark & \cmark \\
 \bottomrule
 \end{tabular}
 }
\end{table*}

%% file: example_shorter.tex
\section{Examples of Translating ARs to MDPs}
\label{sec:example}

We now give two examples of translating an AR problem into an \MDP{}. 
Once modeled as an \MDP{}, we can use various off-the-shelf algorithms (from planning or RL) 
to learn a policy to generate ARs. 

\belowcaptionskip=1pt
\begin{figure}[h!]
    \centering
    \begin{subfigure}[t]{0.70\columnwidth}
        \centering
        \includegraphics[width=\textwidth]{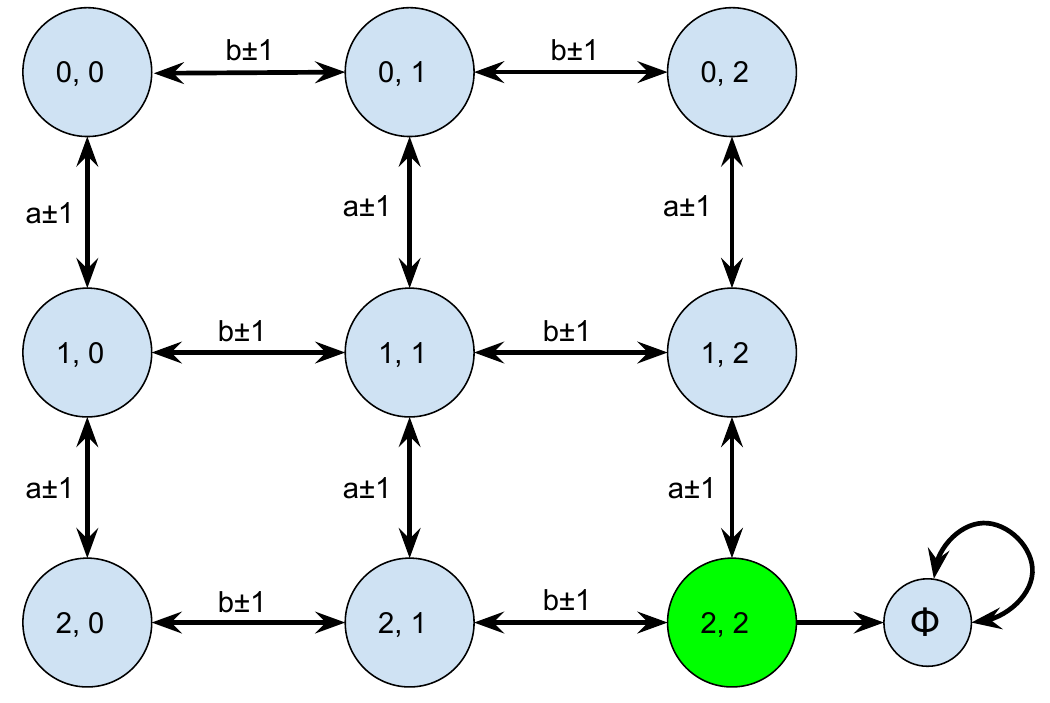}
        \caption{}
        \label{fig:example1}
    \end{subfigure}
    \hfill
    \begin{subfigure}[t]{0.70\columnwidth}
        \centering
        \includegraphics[width=\textwidth]{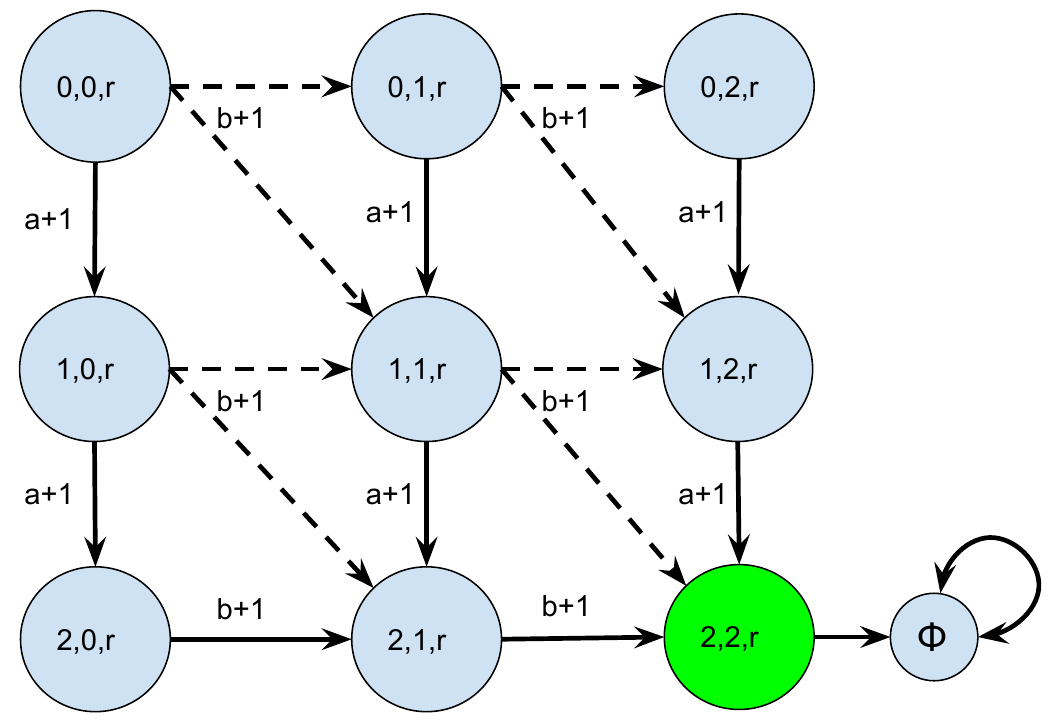}
        \caption{}
        \label{fig:example4}
    \end{subfigure}
   \caption{
   Transition function for the two examples. Circles show all the states, and edges show possible transitions. 
   1) Left-hand-side shows the transition function for a dataset with two features \var{a} and \var{b}, with no restrictions on the values both of them can take within the input domain. The transition edges are therefore bidirectional.  
   2) Right-hand-side shows the transition function for a dataset with three features: age (\var{a}), education-level (\var{b}), and race (\var{r}). 
   The transition edges are unidirectional as both age and education cannot decrease. 
   Since race is immutable, there are no actions for \var{r}. Since an increase in education stochastically affects age, the dashed edges represent a 50\% probability of transition. }
   \label{fig:example14_}
\end{figure}

\textbf{Example 1:} 
Consider two categorical features $\var{a}, \var{b} \in \{0, 1, 2\}$. 
The combinations of possible values for \var{a} and \var{b} form the state space for the \MDP{} (represented by $\mathcal{S}$). 
The directed edges in~\Cref{fig:example1} show that upon taking a specific action, 
an agent can move from one state to another, e.g., it transits from state \var{(0, 1)} 
to \var{(0, 2)} by taking the action \var{b+1}, which increments the value of feature \var{b} by 1. 
Actions \var{a+1} and \var{a-1} respectively increase and decrease the value of feature \var{a} by 1 (similarly for feature \var{b}. 
These actions constitute the action space for the \MDP{} (represented by $\mathcal{A}$). 
The third component of the \MDP{} is the transition function which is 
represented by $\mathit{T: s \times a \to s'}$. 
This denotes that if an agent takes action $a$ in state $s$ then it will move to state $s'$. 
This transition function is deterministic because taking the action $a$ in state $s$ will always land the agent in the state $s'$. 

The final component of the \MDP{} is the reward function. Taking action costs something 
(negative reward), and reaching desirable states generate a positive reward. 
In this \MDP{} taking any action costs a constant amount of 1 and reaching the terminal state 
($\phi$) gives a reward of +10. 
The terminal state ($\phi$) can only be reached via (2,2) (using any action), the state in green color. 
All actions in the terminal state lead to itself with 0 cost. 
This represents the situation in which a ML model classifies only (2,2) in the desired class. 

The aim is to learn a policy that reaches a terminal state from any state at the lowest cost 
(e.g., taking the fewest number of steps). 
Cost (or reward) can be discounted in the traditional way using a discount factor 
$\gamma \in [0,1)$. 
%
%
%
Formally, for this example with a discrete state space and discrete action space, our \MDP{} is:
\begin{itemize}[noitemsep,topsep=0pt,parsep=0pt,partopsep=0pt,leftmargin=*]
    \item States = $\{s \in \mathcal{S}: \{0,0\}, \{0,1\}, \{0,2\}, \{1,0\}, \dots \}$. 
    \item Actions = $\{a \in \mathcal{A}: \var{a+1}, \var{a-1}, \var{b+1}, \var{b-1} \}$. 
    \item Transition function $\mathit{T : \mathcal{S} \times \mathcal{A} \rightarrow \mathcal{S}}$ 
    \item Reward function $r : \mathcal{S} \times \mathcal{A} \to \mathbb{R}$. 
    \item Discount factor $\gamma \in [0,1)$, capturing the tradeoff between current and future reward. 
\end{itemize}
Our goal is finding a policy $\pi: \mathcal{S} \to \mathcal{A}$ that, 
given a state $s \in \mathcal{S}$ (an input datapoint), 
returns an action $a \in \mathcal{A}$ that represents the best first step to take 
to reach a new state, hopefully closer to the ML model's decision boundary.  
\toolname would then call this precomputed policy repeatedly to 
find an optimal path to a counterfactual state. 

\smallskip

\textbf{Example 2:}
Now, consider a more realistic dataset having 3 features: age (denoted by \var{a}), education-level (denoted by \var{b}), and race (denoted by \var{r}). 
This is accompanied by real-world constraints like age and education-level cannot decrease, education-level affects age, and race is immutable. 
When we increase the education-level (\var{b}) by 1, there is a 50\% chance that 
age group (\var{a}) will remain the same and a 50\% chance that it will increase by 1. 
These interactions between features can be captured by a structural causal model (\SCM{}), as we discuss in Section~\ref{sec:algo}. 
The transition function for the \MDP{} representing the AR problem for this dataset is, therefore, stochastic. 

Defined formally, here are the components for this \MDP{}:
\begin{itemize}[noitemsep,topsep=0pt,parsep=0pt,partopsep=0pt,leftmargin=*]
    \item States = $\{s \in \mathcal{S}: \{0,0,0\}, \{0,1,0\}, \{0,2,1\}, \dots \}$. 
    \item Actions = $\{a \in \mathcal{A}: \var{a+1}, \var{a-1}, \var{b+1}, \var{b-1} \}$. 
    \item Transition function $\mathit{T : \mathcal{S} \times \mathcal{A} \times \mathcal{S'} \to}$ \{0,1\} s.t.\ 
    $\forall s \in \mathcal{S}, \forall a \in \mathcal{A},\ \ \sum_{s' \in \mathcal{S}} T(S, A, S') = 1.$
    \item Reward function $r : \mathcal{S} \times \mathcal{A} \to \mathbb{R}$. 
    \item Discount factor $\gamma \in [0,1)$. 
\end{itemize}


\Cref{fig:example4} shows the transition function for this problem. 
The action that increases the education-level (\var{b}) now has a probabilistic transition 
to two destination states, represented by dashed unidirectional edges. 
Each transition edge has a 50\% probability of occurrence. 
Unidirectionality comes from the fact that education-level cannot decrease. 
The edges change feature \var{a} are also unidirectional as age cannot decrease. 
The reward function is identical to the previous example; optionally, it can be changed to accommodate adherence to the data manifold (\Cref{sec:desiderata}) or having different costs for changing different features, which we describe in \Cref{sec:algo}. 
Additional examples can be found in~\Cref{sec:example_remain}. 

%% file: algorithm.tex
\section{An Algorithmic Approach for Generating MDPs from AR Problems}
\label{sec:algo}


\SetKwProg{Fn}{Function}{}{}
\RestyleAlgo{algoruled}
\newcommand\mycommfont[1]{\scriptsize\ttfamily\textcolor{blue}{#1}}
\SetCommentSty{mycommfont}

\newcommand{\xAlCapSty}[1]{\small\sffamily\bfseries\MakeUppercase{#1}}
\SetAlCapSty{xAlCapSty}


\newcommand\mynlfont[1]{\scriptsize\sffamily{#1}}
\SetNlSty{mynlfont}{}{}

\SetSideCommentLeft

\begin{algorithm}[h!]
 \DontPrintSemicolon 
 \SetKwInOut{KwIn}{Input}
 \SetKwInOut{KwOut}{Output}
 \SetKwData{Da}{\textsf{D}}
 \SetKwData{La}{L}
 \SetKwData{Un}{Un}
 \SetKwData{Bin}{Bin}
 \SetKwData{f}{f}
 \SetKwData{SCM}{\textsf{SCM}}
 \SetKwData{K}{K}
 \SetKwData{Af}{\textsf{ActF}}
 \SetKwData{Cf}{\textsf{Cat}}
 \SetKwData{Nf}{\textsf{Num}}

 \SetKwData{CAf}{\textsf{CatA}}
 \SetKwData{NAf}{\textsf{NumA}}

 \SetKwData{Hf}{\textsf{One-hot}}
 \SetKwData{CFReward}{\textsf{Pos}}
 \SetKwData{DF}{\texttt{DistF}}
 \SetKwData{DMF}{\texttt{DistD}}
 \SetKwData{MDP}{\textsf{MDP}}
 \KwIn {Training Dataset (\Da), ML model (\f), 
 Structural Causal Model (\SCM), 
Numerical actionable features (\NAf), Categorical actionable feature (\CAf), 
 Data Manifold distance function (\DMF), 
 Data Manifold adherence ($\lambda$), 
 Desired Label (\La), 
 Distance Function (\DF), Discount Factor ($\gamma$)}
 \KwOut {\MDP}
 
 \caption{\label{alg:mdp} Generate \textsf{MDP} from an Algorithmic Recourse Problem}
 
 \SetKwFunction{TransitionFunction}{Transition}
 \SetKwFunction{RewardFunction}{Reward}
 \SetKwFunction{Allowed}{Allowed}
 \SetKwFunction{InDomain}{InDomain}
 \SetKwFunction{argmax}{argmax}
 \SetKwFunction{Parent}{Parent}
 \SetKwFunction{BinFn}{BinFn}
 \SetKwFunction{cost}{Cost}
 
 \SetKw{break}{break}
 \SetKwData{PMa}{M'}
 \SetKwData{IncNum}{$F_{i}+$}
 \SetKwData{DecNum}{$F_{i}-$}
 \SetKwData{IncCat}{$F_{j}{+}1$}
 \SetKwData{DecCat}{$F_{j}{-}1$}
 \SetKwData{Sa}{\textsf{CurrState}}
 \SetKwData{NSa}{\textsf{State'}}
 \SetKwData{FSa}{\textsf{NextState}}
 \SetKwData{Ac}{$A$}
 \SetKwData{Cost}{Cost1}
 \SetKwData{ManifoldCost}{Cost2}
 \SetKwData{Reward}{\textsf{CFReward}}
 \SetKwData{Ua}{\textsf{U}}
 \SetKwData{Va}{\textsf{V}}
 \SetKwData{NVa}{\textsf{V'}}
 \SetKwData{Fa}{\textsf{F}}
 
 
 \tcp{States consist of all numerical ($\Nf$) and categorical ($\Cf$) features.}
 $\mathit{\text{State space } \mathcal{S} \subseteq \mathbb{R}^{\vert\Nf\vert}\times \mathbb{Z}^{\vert\Cf\vert} }$ \\ \label{line:state}
 
 \tcp{Actions change the actionable numerical and categorical features.}
 $\mathit{\text{Action space } \mathcal{A} \subseteq \mathbb{R}^{\vert\NAf\vert}\times \mathbb{Z}^{\vert\CAf\vert}; \text{denote actions } A \in \mathcal{A} }$  \\ \label{line:action}

 \Fn{$\mathit{\RewardFunction(\f, \La, \Sa, \Ac, \Da, \lambda, \DMF, \SCM)}$}{ \label{line:rewardfunc}
 $\mathit{\FSa \leftarrow \TransitionFunction(\Sa, \Ac, \SCM)}$ \\
 \eIf{$\mathit{\argmax(\f(\FSa))} = \La$}
 {$\mathit{\Reward \leftarrow \CFReward}$  \tcp{High positive reward}}
 {$\mathit{\Reward \leftarrow \f(\FSa)[\La]}$ \tcp{Probability of classification in the desired class} \label{line:otherreward}}
 \Return $\DF(\Sa, \Ac, \Da)$ \tcp{action cost} \label{line:reward1}
 $+ \lambda * \DMF(\FSa, \Da)$  \tcp{Manifold distance cost} \label{line:reward2}
 $+\Reward$ \tcp{Counterfactual label reward} \label{line:CFreward}
 }
 
 
\Fn{$\mathit{\TransitionFunction(\Sa, \Ac, \SCM})$ }{ \label{line:transitfunc}
    \tcp{Action does not violate feature domain and unary constraints}
    \eIf{\Allowed(\Ac) \& \InDomain(\Ac) \label{line:updated}}
    {$\mathit{\FSa \leftarrow \Sa + \Ac}$ \tcp{Modify features} \label{line:change}}
    {\Return \Sa \label{line:illegalaction}}
    
    \tcp{Modify the endogenous features}
    
    \For{$\mathit{\Va \in \SCM}$ \label{line:exogenous}} {
    \uIf{$\mathit{\Ac \in \Parent(\Va)}$ }
    { 
        $\mathit{\FSa[\Va] \leftarrow \Fa(\Ua)}$ 
        \tcp{Stochastic or deterministic update of endogenous features} \label{line:endoupdate}
    }
    }
 \Return \FSa
}
 
$\mathit{\MDP \leftarrow \{\mathcal{S}, \mathcal{A}, \TransitionFunction, \RewardFunction, \gamma\}}$
\end{algorithm}

We now present a general approach for translating an AR problem setup into an \MDP{}. 
\Cref{alg:mdp} generates all components of an \MDP{}: state space, action space, 
transition function, reward function, and additional parameters such as discount factor. 
We detail this process below. 

\textbf{State space. }
Features can be broadly categorized into numerical (\textsf{Num}) and categorical (\textsf{Cat}) kinds. 
Numerical features can take real number values within a specified domain, while 
categorical features are mapped to a set of integers. 
Consequently, the state space $\mathcal{S}$ of our \textsf{MDP} (\cref{line:state}) 
consists of the product of the continuous domains for numerical features 
(a subset of $\mathit{\mathbb{R}^{\vert \textsf{Num}\vert}}$) and product of the 
integer domains for categorical features 
(a subset of $\mathit{\mathbb{Z}^{\vert \textsf{Cat}\vert}}$). 

\textbf{Action space. }
To facilitate capturing actionability~\citep{Ustun19:Actionable} and causal relationships between features~\citep{karimi_algorithmic_2020}, we further categorize features as follows: 
\begin{itemize}[noitemsep,topsep=0pt,parsep=0pt,partopsep=0pt,leftmargin=*]
    \item \emph{Actionable} features can be directly changed by an individual, e.g., income, education level, age. 
    \item \emph{Mutable but not actionable} features are mutable but cannot be modified directly by an individual, 
    e.g., credit score cannot be directly changed by a person, it changes due to change in features like income and credit history. 
    \item \emph{Immutable} features cannot change, e.g., race, birthplace. 
\end{itemize}
The agent is permitted to change only the actionable numerical and categorical features (denoted by \textsf{NumA} and \textsf{CatA}). 
Consequently, the action space $\mathcal{A}$ is a subset of 
$\mathbb{R}^{\vert\textsf{NumA}\vert} \times \mathbb{Z}^{\vert\textsf{CatA}\vert}$ (\cref{line:action}). 
Categorical features are changed within their discrete domain, 
while numerical features are changed within their continuous domain. 
\Cref{line:updated} further enforces the infeasibility of out-of-domain actions. 


\textbf{Transition function.}
The transition function (\cref{line:transitfunc}) finds the modified state when an action is taken. 
This function is influenced by the structural causal model (\textsf{SCM}), which is an 
{\it optional} input to \Cref{alg:mdp}. 
An \textsf{SCM} consists of a triplet \textsf{M} = $\langle \textsf{U}, \textsf{V}, \textsf{F}\rangle$. 
$\textsf{U}$ is the set of \emph{exogenous} features and $\textsf{V}$ is the set of \emph{endogenous} features. 
In terms of a causal graph, the exogenous features $\textsf{U}$ consist of 
features that have no parents, i.e., they can change independently. 
The endogenous features $\textsf{V}$ consists of features that have parents in $\textsf{U}$ 
and/or other features in $\textsf{V}$. They change as an effect of change in their parents. 
$\textsf{F}$ is the set of functions that determine the relationship between exogenous and endogenous features. 
They are termed as structural equations. 

Since knowing the exact \textsf{SCM} is often infeasible, 
\citet{mahajan_preserving_2020} overcome this limitation by utlizing constraints from domain knowledge. 
\Cref{alg:mdp} also accepts such contraints in unary (\textsf{Un}) and binary (\textsf{Bin}) forms.  
Even if this does not provide us with the precise functional form of the constraint, 
its nature can help the \toolname's recourses to be realistic. 
Unary constraints are derived from the property of one feature, e.g., 
age and education level cannot decrease. 
Binary constraints are derived from the relation between two features, e.g., 
if the education level increases, age increases. 
If an action does not violate the domain of the feature it is changing, 
nor the constraints in the \textsf{SCM}, then the feature is modified in 
\textsf{NextState} (\cref{line:change}). 
%
If the modified feature is an exogenous feature, we update its children 
using the \textsf{F} functions (\cref{line:exogenous}-\ref{line:endoupdate}). 

Note that if no SCM is input to the algorithm, that will allow transitions 
from any state to any state (with intermediate states), and 
\toolname would generate ARs using this unconstrained transition function.


\textbf{Reward function. }
\Cref{line:rewardfunc} defines a reward function that, given a state and an action, 
returns a reward based on three components derived from the initial AR problem: 
\begin{itemize}[noitemsep,topsep=0pt,parsep=0pt,partopsep=0pt,leftmargin=*]
 \item Given the current state (\textsf{CurrState}), action (\textsf{$A$}), 
 training dataset (\textsf{D}), and distance function \texttt{DistF}, 
 the first part returns the appropriate cost to take that action (\cref{line:reward1}). 
 The distance function can either be the $\ell_p$ norm of the change produced by the action or a more complex function. 
 
 \item The second part adds a cost if a datapoint is far from the training data manifold 
 (\cref{line:reward2}) (which is computed using the \texttt{DistD} function)
 A $\lambda$ factor is used to control the strictness of data manifold adherence. 
 
 \item The third part rewards the agent with a large positive value if a 
 counterfactual state is reached (\textsf{CFReward} in \cref{line:CFreward}). 
 To avoid sparse rewards, we partially reward the agent with a small reward equal to the probability of \textsf{NextState} being classified in the desired class (\cref{line:otherreward}). 
 However, the sparse rewards can only be used if the underlying ML model provides the class label probabilities instead of only the class label, e.g., a neural network or random forest. 
 
\end{itemize}

\textbf{Other parameters.}
MDPs require additional parameters such as the discount factor $\gamma \in [0,1]$. 
At a high level, setting $\gamma < 1$ penalizes longer paths; 
for additional intuition, see~\citet{RLBook}. 
We note that $\lambda$, \textsf{DistD}, and \textsf{DistF} are user-specified and 
domain-specific parameters that directly impact the reward function for the \textsf{MDP}. 
We instantiate them in the evaluation section (see~\cref{sec:eval}). 

%% file: evaluation.tex
\section{Evaluation}
\label{sec:eval}

\setlength{\belowcaptionskip}{-9pt}
We provide experimental validation of \toolname using three real-world datasets and comparison using nine baselines. 
Our research questions (RQ) are motivated by the recourse desiderata discussed in \Cref{sec:desiderata}, 
and are enumerated here:

\begin{description}[noitemsep,topsep=-1pt,parsep=0pt,partopsep=0pt,leftmargin=*] 
 \item[\RQ{1}] Does \toolname successfully generate ARs for various input datapoints (validity)? 
 \item[\RQ{2}] How much change is required to reach a counterfactual state (proximity)? 
 \item[\RQ{3}] How many features are changed to reach a counterfactual state (sparsity)? 
 \item[\RQ{4}] Do the generated ARs adhere to the data manifold (realisticness)? 
 \item[\RQ{5}] Do the generated ARs respect causal and feature immutability constraints (feasibility)? 
 \item[\RQ{6}] How much time does \toolname take to generate ARs (amortizability)? 
\end{description}




\vspace{-4pt}
\paragraph{Datasets. }
Motivated by most previous AR generating approaches~\cite{verma2020CFsurvey}, 
we use three datasets in our experiments: German Credit, Adult Income, and Credit Default~\citep{UCI-repo}. 
These datasets have 20, 13 (omitted \var{education-num} as it has one to one mapping 
with \var{education}), and 23 features respectively. 
We split the datasets into 80\%-10\%-10\% for training, validation, and testing, respectively. 
Each dataset has two labels, `1' and `0', where `1' is the desired label. 
We trained a simple classifier: a neural network with two 
hidden layers (5 and 3 neurons) with ReLU activations. 
The test accuracy of the classifier was 83.0\% for German Credit, 83.7\% for Adult Income, and 83.2\% for Credit Default. 
Note that the classifier's accuracy is relatively 
less important for \toolname's validation. 

\vspace{-4pt}
\paragraph{Implementation Algorithm. }
Any appropriate method for computing an optimal policy $\pi^*: \mathcal{S} \to \mathcal{A}$, 
or any approximately optimal policy, to the \MDP{} output of Algorithm~\ref{alg:mdp} can be used. 
Our \MDP has a continuous state and action space, and therefore we use a policy gradient algorithm. 
Specifically, we use proximal policy optimization (PPO) with generalized advantage estimate (GAE)~\citep{PPO2017,a3c-mniha16,GAE2018} to train the agent. 
We justify this choice and answer several other related questions in~\Cref{sec:justification}. 
The features in all datasets are scaled between $-1$ and $1$ before training both the 
ML model and the RL agent. 


\subsection{Baselines} Since, to our knowledge, \toolname is the first approach to generate 
amortized ARs for black-box models, there exist no previous approaches which we can directly 
compare against. Nevertheless, we compare \toolname to several previous popular 
AR generating approaches. 

\noindent\textbf{Baselines we developed.} To compare \toolname to approaches that 
generated ARs in an amortized manner for black-box models, we developed two baselines: 
\begin{itemize}[noitemsep,topsep=0pt,parsep=0pt,partopsep=0pt,leftmargin=*]
 \item \textbf{Random:} This approach tries to reach a counterfactual state by executing 
 random actions from the action space. 
 
 \item \textbf{Greedy:} At each step, this approach greedily chooses the action (among all actions) which gives 
 the highest reward. 
\end{itemize}

\noindent\textbf{Previous AR generating approaches.}
Based on the level of required model access, AR generating approaches can be categorized as: 
1) access to complete model internals, i.e., weights of neurons or nodes of decision trees, 
2) access to model gradients (restricted to differentiable models like neural networks), and 
3) access to only the \functionname{predict} function (black-box). 
We choose popular methods from all categories:
\begin{itemize}[noitemsep,topsep=0pt,parsep=0pt,partopsep=0pt,leftmargin=*]
\item \textbf{Complete model internal access.} 
We chose MACE~\cite{karimi_model-agnostic_2020} from this category. 

\item \textbf{Gradients access.} Here we chose DiCE-Gradient \cite{mothilal_explaining_2020} 
and DiCE-VAE \cite{mahajan_preserving_2020}. 
Notably, DiCE-VAE is the {\it only other} amortized AR generation method, 
however, it requires gradients and is restricted to differentiable models. 

\item \textbf{Black-box.} Open-source repository of the aforementioned DiCE method 
also had three black-box and model-agnostic approaches, namely: 
DiCE-Genetic, DiCE-KD-Tree, and DiCE-Random. 
We choose these three and Prototypes \citep{van_looveren_interpretable_2020} for this category. 
We did not compare with MOC \citep{dandl_multi-objective_2020} as DiCE-Genetic it also a 
genetic algorithm based approach and has uses Python code. 

\end{itemize}


\subsection{Experimental Methodology}

\begin{table}   
    \footnotesize
    \centering
    \caption{Causal constraints and immutable features for the datasets. 
    We assume \toolname is provided with them. 
    }
    \label{tab:causal-rels}
    \resizebox{\columnwidth}{!}{
    \begin{tabular}{m{0.15\columnwidth}m{0.45\columnwidth}m{0.4\columnwidth}}
        \toprule
        Dataset & Causal constraints & Immutable features \\
        \midrule
        German Credit & Age and Job cannot decrease  & Foreign worker, Number of liable people, Personal status, Purpose \\ \\
        
        Adult \mbox{Income} & Age and Education cannot decrease, increasing Education increases Age & Marital-status, Race, Native-country, Sex \\ \\
        
        Credit \mbox{Default} & Age and Education cannot decrease, increasing Education increases Age & Sex, Marital status \\
        \bottomrule
    \end{tabular}
    }
\end{table}

Here we describe the specific details of some approaches:

\vspace{-7pt}

\paragraph{\toolname specifics.}
As stated in~\Cref{sec:algo}, the recourses generated by \toolname are realistic if provided 
with the actionability of features and causal constraints. 
These constraints can be provided using the complete/partial SCM 
of the data generating process or using domain knowledge. 
We assume these constraints are provided to \toolname and are shown in~\Cref{tab:causal-rels}. 
As described in Algorithm~\ref{alg:mdp}, this directly impacts the transition function. 
We use a particular instantiation of~\Cref{alg:mdp} in the experiments:
\begin{itemize}[noitemsep,topsep=0pt,parsep=0pt,partopsep=0pt,leftmargin=*] 
 \item Action space: To produce sequential ARs, actions modify only one feature at a time. 
 However, endogenous features may simultaneously change due to change in their parent. 
 \item Cost of action: We treat \textsf{DistF} function as a hyperparameter and 
 use several values for it in the experiments. 
 \item Data manifold distance: Following previous work~\citep{dandl_multi-objective_2020,Kanamori2020:DACE}, 
 we train $k$-Nearest Neighbor (KNN) algorithm on the training dataset 
 and use it to find the $\ell_1$ distance of a given datapoint from its 
 nearest neighbor ($k = 1$) in the dataset (\textsf{DistD}). 
 We use several values of the adherence factor $\lambda$ in the experiments. 
 \item Counterfactual state reward (\textsf{CFReward}): The agent receives a 
 reward equal to the probability of its state belonging to the desired class (this ranges between 0 and 1). 
 However, when a counterfactual state is reached, the agent is rewarded with 100 points. 
 \item Discount Factor: We use a discount factor $\gamma=0.99$. 
 This value encourages the agent to learn a policy that takes a few steps to reach a counterfactual state. 
\end{itemize}
We explore the impact of $\lambda$ hyperparameter in~\Cref{sec:hyperparam}, 
and give more implementation details in~\Cref{sec:implementation-details}. 
\vspace{-6pt}
\paragraph{MACE specifics.} MACE requires as input the type of ML classifier to be used. 
We could not use a neural network because of the MACE's long runtime (see \cref{sec:results}), and 
and therefore choose logistic regression (LR) and random forest (RF), which had a reasonable runtime.

All approaches are requested to generate ARs for the test datapoints that are 
predicted as `0' by the classifier. 
Due to the small size of the German Credit dataset, we generate ARs for datapoints 
that are predicted as `0' both in the training and test sets. 
Thus we request ARs for 257 datapoints in the German credit, 7229 datapoints in the 
Adult Income, and 5363 datapoints in the Credit Default datasets. 
Since MACE uses a different classifier, the number of datapoints predicted as `0' were slightly different. More details are provided in~\cref{sec:results}. 
\toolname, random, and greedy approaches stop when they reach a counterfactual state 
(predicted as `1') or exhaust 50 actions. Other baselines have no such timeout.

\subsection{Results}
\label{sec:results}

\input{results_table}

\Cref{tab:finalresults} shows the performance of \toolname and all the baselines on the 
recourse desiderata. We report the average validity, average proximity 
(separately for the numerical and categorical features), average sparsity, 
average data manifold distance, average causal constraints adherence, 
and the average time to generate the ARs per datapoint. 

\noindent \textbf{Answer to \RQ{1}:} As shown in \Cref{tab:finalresults}, 
\toolname has very high validity for all datasets. 
For Adult Income, \toolname gets the highest validity at 100\%, 
while for Credit Default and German Credit, it achieves the second and third highest validity, respectively. 
Random and greedy approaches have low validity in general. 
DiCE-Genetic has validity in the high range, but this comes at the cost of proximity, sparsity, and data manifold distance. 
DiCE-KDTree is unable to generate AR even for a single datapoint in all three datasets. 
DiCE-Random achieves 100\% validity for all datasets, and just like DiCE-Genetic, this comes at the cost of proximity, sparsity, and data manifold distance. 
The conclusion is similar for DiCE-Gradient's and Prototypes' validity. 
DiCE-VAE's validity is lower than 80\% for all datasets. 
MACE also achieves 100\% validity but is very expensive to run. 
Due to this, it was impractical to run MACE for the larger datasets, 
Adult Income and Credit Default (we show MACE run only for the German Credit dataset). 
MACE was even more expensive when the underlying classifier was a neural network, 
and we had to abandon that experiment. 
For the classifiers used for MACE, `0' was predicted for 210 datapoints by 
logistic regression (LR) and 287 datapoints by random forest (RF). 
MACE was supposed to generate ARs for these datapoints.




\textbf{Answer to \RQ{2}:} We measure proximity for numerical and categorical features separately 
(Prox-Num and Prox-Cat, respectively). 
For numerical features, the distance is the sum of the $\ell_1$ norm respectively divided by 
the median average deviation for each numerical feature. 
For categorical features, the distance is the number of categorical features changed 
divided by the total number of categorical features. 
These metrics were proposed and used in previous works~\cite{mahajan_preserving_2020}. 
\toolname's ARs are most proximal for Adult Income and Credit Default datasets, 
and second best for German Credit. 
The random approach, Prototypes, and the five variants of DiCE have large proximity values. 
The greedy approach performs well on this metric, but its validity is very low. 
MACE's performance is about average. 



\noindent \textbf{Answer to \RQ{3}:}
\toolname achieves the lowest sparsity among all approaches. 
Following previous works~\cite{mothilal_explaining_2020}, 
we measure sparsity at the start and endpoint of a recourse. 
Random, Prototypes, DiCE-VAE, DiCE-Genetic, and DiCE-Gradient's performance is abysmal. 
This is surprising because DiCE-Gradient has a post-hoc step 
specifically for reducing sparsity. 
Greedy, MACE, and DiCE-Random's performance is about average. 


\noindent \textbf{Answer to \RQ{4}:} 
\toolname achieves low average manifold distance. 
It performs second best for Adult Income and Credit Default and is in the middle for German Credit. 
The greedy approach, MACE, and DiCE-VAE also perform well on this metric. 
The random approach, Prototypes, and all variants of DiCE (except DiCE-VAE) perform poorly on this metric. 




\noindent \textbf{Answer to \RQ{5}:} 
By construction, \toolname always respects causal constraints encoded in 
its transition function: it has 100\% adherence in all datasets. 
The DiCE based approaches (except DiCE-VAE), MACE, and random approach take as input the immutable features, 
but not the other causal constraints and hence do not perform well. 
DiCE-VAE and Prototypes do not accept immutable features and 
hence perform the worst for this metric. 
The greedy approach performs well on this metric, even though it does not have a 
knowledge of the causal constraints. 




\noindent \textbf{Answer to \RQ{6}:} 
The final column in \Cref{tab:finalresults} reports the average computation time per AR. 
Owing to amortization, \toolname can generate ARs very quickly and 
takes the lowest time among all approaches. 
The next best performers are DiCE-VAE and DICE-Random. 
\toolname is \textbf{2$\times$} faster than DiCE-VAE on average 
(up to \textbf{8$\times$} faster), \textbf{8$\times$} faster than DiCE-random on average 
(up to \textbf{15$\times$} faster). DiCE-random and random approach perform similarly. 
The difference even more staggering for DiCE-Genetic, Prototypes, and greedy approach. 
MACE and Dice-Gradient were the slowest. 
\toolname is about \textbf{1000$\times$} faster than MACE on average 
(up to \textbf{1447$\times$} faster) and \textbf{4500$\times$} faster than DiCE-Gradient 
on average (up to \textbf{9400$\times$} faster). 
While amortization allows for the rapid generation of new ARs, 
there exists a one-time training cost. We give details about it in~\Cref{sec:implementation-details}. 



%% file: results_table.tex
\begin{table*}[t]
\centering
\caption{Comparison of \toolname to all baselines for various AR evaluation metrics. 
    \texttt{Validity} is the percentage an AR is actually classified in the desired class. 
    \texttt{Prox-Num} and \texttt{Prox-Cat} refers to the L1 distance of the datapoint and its AR for the numerical and categorical features respectively. 
    \texttt{Sparsity} is the number of features that were changed to produce the AR. 
    \texttt{Manifold dist.} is the distance of the AR as returned by the trained kNN algorithm. 
    \texttt{Constraints} refer to the causal constraints adherence by the generated AR. 
    \texttt{Time} is the average time to generate ARs. 
    For \texttt{Validity} and \texttt{Constraints}, a higher value is better, and for all other columns, 
    a lower value is better. 
    MACE and DiCE-Gradient could not be run for larger datasets owing to their large computation time. 
    }
    
\resizebox{\textwidth}{!}{
\begin{tabular}{l l l l l l l l l l}
\toprule
Dataset  & Approach & \#DataPts. & Validity & Prox-Num & Prox-Cat & Sparsity & Manifold dist. & Constraints & Time (s) \\
\midrule
\multirow{10}{*}{\rotatebox[origin=c]{90}{German Credit}}
    &   Random        & 257 & 23.7      & 0.17     & 0.57      & 11.33     & 1.08       & 41.0       & 0.31    \\ 
    &   Greedy        & 257 & 49.8      & \x{0.07} & 0.087     & 1.81      & 0.48       & \x{100.0}  & 4.59    \\ 
    &   DiCE-Genetic  & 257 & 98.1      & 0.67     & 0.26      & 6.52      & 2.39       & 45.6       & 1.71    \\ 
    &   DiCE-KDTree   & 257 & 0.0       & N/A      & N/A       & N/A       & N/A        & N/A        & 0.17    \\ 
    &   DiCE-Random   & 257 & \x{100.0} & 0.33     & 0.10      & 1.93      & 2.40       & 93.4       & 0.17    \\ 
    &   Prorotypes    & 207 & \x{100.0} & 0.26     & 0.58      & 13.1      & 1.0        & 5.3        & 25.9     \\  
    &   DiCE-Gradient & 257 & \x{100.0} & 0.27     & 0.29      & 6.33      & 2.19       & 82.9       & 7.10    \\ 
    &   DiCE-VAE      & 257 & 77.8      & 0.80     & 0.42      & 10.12     & 0.97       & 5.0        & 0.15    \\      
    &   MACE (LR)     & 210 & \x{100.0} & 0.36     & \x{0.017} & 1.99      & 0.60       & 97.1       & 38.45   \\ 
    &   MACE (RF)     & 287 & \x{100.0} & 0.22     & 0.02      & 2.64      & \x{0.38}   & 74.2       & 101.29  \\    
    &   \toolname     & 257 & 97.3      & 0.10     & 0.063     & \x{1.22}  & 0.72       & \x{100.0}  & \x{0.07}    \\ 
\midrule
\multirow{8}{*}{\rotatebox[origin=c]{90}{Adult Income}}
    &   Random        & 7229 & 80.9      & 0.56     & 0.77     & 10.07    & 1.00     & 29.0        &  0.25    \\  
    &   Greedy        & 7229 & 97.7      & \x{0.04} & 0.02     & 1.18     & \x{0.17} & 95.0        &  0.27    \\  
    &   DiCE-Genetic  & 7229 & 89.5      & 0.71     & 0.27     & 4.43     & 0.46     & 23.0        &  3.43    \\  
    &   DiCE-KDTree   & 7229 & 0.0       & N/A      & N/A      & N/A      & N/A      & N/A         &  0.59    \\  
    &   DiCE-Random   & 7229 & \x{100.0} & 0.82     & 0.04     & 1.64     & 1.24     & 90.0        &  0.22    \\  
    &   Prototypes    & 500  & \x{100.0} & 0.29     & 0.57     & 9.0      & 0.68     & 22.8        &  28.9      \\  
    &   DiCE-Gradient & 500  & 84.0      & 0.18     & 0.012    & 2.78     & 0.51     & 82.4        &  59.75   \\  
    &   DiCE-VAE      & 7229 & 77.1      & 0.75     & 0.65     & 9.99     & 0.30     & 0.13        &  0.12    \\  
    &   \toolname     & 7229 & \x{100.0} & \x{0.04} & \x{0.0}  & \x{1.00} & 0.18     & \x{100.0}   &  \x{0.015}  \\  
\midrule
\multirow{8}{*}{\rotatebox[origin=c]{90}{Credit Default}}
    &   Random        &  5363 & 12.8      & 4.85     & 0.68      & 14.54      & 1.30      & 41.5     &  0.63    \\   
    &   Greedy        &  5363 & 65.1      & 0.15     & \x{0.072} & 1.25       & \x{0.22}  & 99.9     &  4.67    \\   
    &   DiCE-Genetic  &  5363 & 92.6      & 3.93     & 0.49      & 16.67      & 2.75      & 27.9     &  3.58    \\   
    &   DiCE-KDTree   &  5363 & 0.0       & N/A      & N/A       & N/A        & N/A       & N/A      &  0.45    \\   
    &   DiCE-Random   &  5363 & \x{100.0} & 5.80     & 0.20      & 2.33       & 3.09      & 97.7     &  0.39    \\   
    &   Prototypes    &  500  & \x{100.0} & 4.9      & 0.86      & 21.0       & 1.24      & 0.0      &  27.3    \\ 
    &   DiCE-Gradient &  100  & 81.0      & 0.77     & 0.40      & 15.98      & 1.35      & 85.2     &  479.17  \\   
    &   DiCE-VAE      &  5363 & 76.4      & 1.6      & 0.68      &  20.1      & 0.31      & 8.9      &  0.18    \\  
    &   \toolname     &  5363 & 99.9      & \x{0.01} & 0.11      & \x{1.008}  & 0.32      & \x{100.0} &  \x{0.051}  \\   
    
\bottomrule
    \end{tabular}
    }

    \label{tab:finalresults}
\end{table*}


%% file: conclusion.tex
\section{Conclusion}
\label{sec:conclusion}

%
We propose a novel RL-based approach, \toolname, that generates amortized and 
sequential recourses for black-box ML models. 
To the best of our knowledge, we are the first to propose such an approach. 
The ARs generated by \toolname possess desirable properties and when evaluated on 
the recourse metrics, they perform better than several popular baselines. 


%% file: appendix.tex
\input{background}
\input{related}
\input{example_remaining}
\input{counter_vs_contrastive}
\input{algorithm_justification}
\input{implementation_details}
\input{hyperparam}

%% file: background.tex
\section{Background}
\label{sec:background}
This section provides background about the social implications of ML models and techniques to address concerns, along with a brief introduction to Reinforcement Learning. 

\subsection{Fairness, Accountability, and Transparency of AI and ML}
Fairness and explainability of an ML model are two major themes in the broad area of equitable ML learning research. 

Fairness research mostly proposes algorithms that learn a model that does not discriminate against individuals belonging to disadvantaged demographic groups. Other possibilities of intervention lie in modifying the training data itself. 
Demographic groups are determined by values of sensitive attributes prescribed by law, e.g., race, sex, religion, or the nation of origin.  
ML models can get biased against certain demographic groups because of the bias in their training data, specifically label bias and selection bias. 
Label bias occurs due to manual biased labeling of datapoints belonging to a demographic group, e.g., if individuals from the black community were denied loans in the past irrespective of their ability to pay back, this gets captured in the data from which the model can learn. 
Selection bias occurs when specific subsets of a demographic group are selected, which captures potentially correlations between the prediction target and a specific demographic group, e.g., selecting only defaulters from a demographic group in the training data. 
More than 20 definitions of fairness of an ML model have been proposed in literature~\citep{verma_fairness}. 
\citet{dunkelau_fairness-aware} summarize some of the significant research advances that have been made in fairness research, and is a comprehensive introductory text for understanding the categorization and direction of research.

Explainability research can be broadly divided into model explanation and outcome explanation research problems~\citep{xai-survey4}. The model explanation problem seeks to search for an inherently interpretable and transparent model with high fidelity to the original complex model. Linear models, decision trees, and rule sets are examples of inherently interpretable models. 
There exists techniques to explain complex models like neural networks and tree ensembles using interpretable surrogate like decision tree~\cite{craven_exp1,KRISHNAN_exp2,Chipman_makingsense_exp3,Pedro_exp4} and rule sets~\cite{Deng_exp5,Andrews_exp6}. 
There also exist approaches that can be applied to black-box models~\citep{Andreas_exp7,Krishnan_exp8,Zien_exp9}. 

The outcome explanation problem seeks to find, for a single datapoint and prediction from a model, an explanation of why the model made its prediction. The explanation is either provided in the form of the importance of each feature in the datapoint, or the form of example datapoints.  The first class of methods are called feature attribution methods and are grouped into model-specific~\citep{Khosla_cam,grad-cam} and model-agnostic~\citep{Poulin_explaind,ribeiro_why_2016,Turner2016_MES} kinds.  Example-based approaches return a few datapoints that either have the same class label as the original datapoint or a different class label. 
The motivation for the first is to provide a set of datapoints that must be similar in the input space. 
The motivation for the second is to provide a set of datapoints that serves as a target to achieve in case the individual wants to receive the alternative label. 
The second set of datapoints can be referred to as  \emph{counterfactual explanations}. 

Counterfactual explanations are applicable to supervised machine learning where the desired label has not been obtained for a datapoint. 
Most research in counterfactual explanations assumes a classification setting. 
Supervised ML setup consists of several labeled datapoints, which are inputs to the algorithm, and the aim is to learn a function mapping from the input datapoints (with say m features) to labels. 
In classification, the labels are discrete values. 
The input space is denoted by $\mathcal{X}^m$ and the output space is denoted by $\mathcal{Y}$. 
The learned function is the mapping $f: \mathcal{X}^m \to \mathcal{Y}$ is used to make predictions. 
We expound on counterfactual explanations and their desirable properties in \Cref{sec:desiderata}. 

Major beneficiaries of explainable machine learning include the healthcare and finance sectors, which have a huge social impact~\citep{tjoa2019survey1}. 
We point the readers to surveys in the area of explainable machine learning~\citep{xai-survey2,carvalho2019:survey3,xai-survey4}.


\subsection{Reinforcement Learning}
Reinforcement Learning (RL) is one of the three broad classes of machine learning, along with supervised and unsupervised learning. 
In RL, the goal is to explore a given environment and to learn a policy over time that dictates what action should be taken at a given state. The exploration happens with the help of an agent. 
Therefore, a policy is a mapping from a state to an action. 
When an action is taken at a state, the environment returns with the new state and a reward. 
A good policy aims to maximize the reward over time. 
The calculation of the new state is facilitated through the transition function, whereas the calculation of the reward is done using the reward function. 
Naturally, the agent can either learn policies that are greedy and only focus on immediate reward or learn policies that focus on reward in the long-term. This trade-off is controlled by a discount factor called $\gamma$, whose value lies between 0 and 1 (inclusive of 0 and 1). 
States can either be discrete or continuous. Similarly, actions can also be either discrete or continuous. 
An RL problem is expressed in terms of a Markov Decision Process (MDP), which has five components. We illustrate each of them using the game of chess. 
\begin{itemize}
    \item State space $\mathcal{S}$, which are states an agent might explore. In chess, these are the 64 squares that an agent can move to. 
    \item Action space $\mathcal{A}$, which are the possible actions an agent can take. These might be restricted based on the current state. In chess, the actions depend on the game pieces like a king, queen or pawns, and the given position on the chessboard. The action space is the union of all possible actions. 
    \item Transition function \texttt{T} which given the current state and action, find the new state that the agent will transition to, e.g., moving the pawn by 1 unit north end up putting the agent in the state that is one unit north of its current state. Transition functions can be deterministic or stochastic (see \Cref{sec:example}). 
    \item Reward function \texttt{R} passes the reward to the agent given the action, the current state, and the new state. This reward signal is the main factor that the agent uses to learn a good policy, e.g., winning a game would pass a positive reward, and losing the game would send a negative reward to the agent. 
    \item Discount factor $\gamma$ is associated with the nature of the problem at hand. This is used to decide the trade-off between immediate and long-term rewards. 
\end{itemize}
Many algorithms have been developed to efficiently learn an agent, given the environment like value iteration, policy iteration, policy gradient, actor-critic methods~\cite{RLBook}. 


%% file: related.tex
\section{Related Works}
\label{sec:related}

Literature in counterfactual explanations for ML is relatively recent, with the first proposed algorithm in~\citeyear{wachter_counterfactual_2017}. 
\citet{wachter_counterfactual_2017} proposed finding counterfactuals as a constrained optimization problem where the goal is to find the minimum change in the features such that the new datapoint has the desired label. This approach was gradient-based, did not consider actionability among features, did not adhere to data manifold or respect causal relations, and the optimization problem needed to be solved for generating a CFE for each input datapoint. 
Other desiderata mentioned in~\Cref{sec:desiderata} were proposed by other papers: 1) approaches that generate multiple, diverse counterfactuals for a single input datapoint~\citep{mothilal_explaining_2020,dandl_multi-objective_2020,mahajan_preserving_2020,karimi_model-agnostic_2020,sharma_certifai_2019,russell_efficient_2019}, 2) approaches that generate counterfactual for black-box models and are model-agnostic~\citep{inverse-classification2,medina_comparison-based_2018,guidotti_local_2018,grath_interpretable_2018,sharma_certifai_2019,rathi-generating:2019,white_measurable_2019,poyiadzi_face_2020,keane2020good,dandl_multi-objective_2020}, 3) approaches that generate CFEs adhering to data manifold~\citep{dhurandhar_explanations_2018,dhurandhar_model_2019,joshi_towards_2019,van_looveren_interpretable_2020,mahajan_preserving_2020,pawelczyk_learning_2020,keane2020good,Grace:2019,dandl_multi-objective_2020,Kanamori2020:DACE}, 4) approaches that generate CFEs that respect causal relations~\citep{mahajan_preserving_2020,karimi_algorithmic_2020,karimi-imperfect:2020}, 5) approaches that generate amortized CFEs~\citep{mahajan_preserving_2020}. 

\citet{mahajan_preserving_2020} was the first to propose an approach that can generate multiple CFEs for many datapoints, after optimizing once, therefore \emph{amortized} CFEs, but their approach is gradient-based and therefore works only for differentiable models and it not black-box. 
Our approach overcomes this limitation and generates both \emph{amortized} and \emph{model-agnostic} CFEs, which adhere to data manifold and respect causal relations. 
Out of the previous approaches that respect causal relations, only \citet{mahajan_preserving_2020} works with partial \textsf{SCM}, while others require complete causal graph or complete \textsf{SCM}~\citep{karimi_algorithmic_2020,karimi-imperfect:2020}, which are mostly unavailable in the real world. Our approach also works with a partial \textsf{SCM}. 

All the previous works give a single-shot solution for getting to a counterfactual state from an input datapoint. Our approach overcomes this limitation by proposing a novel algorithm that generates \emph{sequential} CFEs.

\citet{verma2020CFsurvey} and \citet{karimi2020survey} have collected and summarized recent works in counterfactual explainability. We point the readers to these surveys for an excellent in-depth review of the research landscape in this area. 

%% file: example_remaining.tex
\section{Illustrative examples}
\label{sec:example_remain}

This section gives the remaining examples of translating a CFE problem into an \textsf{MDP}. 

\begin{figure}
    \centering
    \includegraphics[width=0.8\columnwidth]{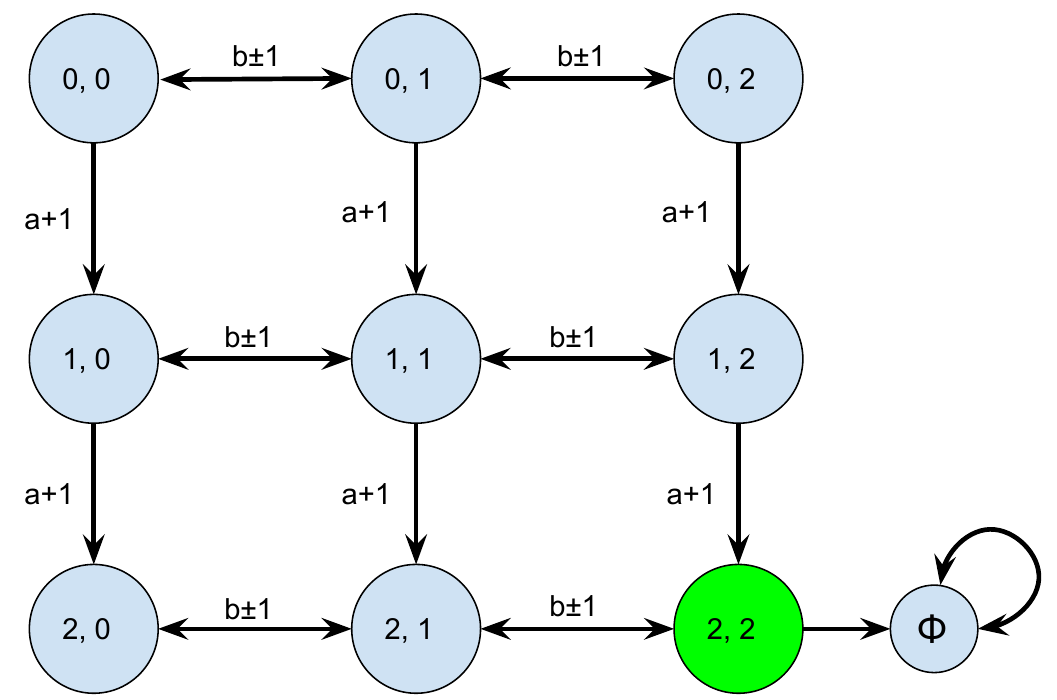}
    \caption{Transition function for a dataset with 2 features, out of which one is age (denoted by \var{a}). Circles show the states and edges show possible transitions. The edges which denote action on the feature \var{a} are unidirectional as age cannot decrease. Action \var{a-1} taken at any state would loop back to the same state. Each action has a constant cost of 1. 
    }
    \vspace{1em}
    \label{fig:example2}
\end{figure}

\textbf{Example 1:}
Let us now consider the example where one of the two features is age (denoted by feature \var{a}). 
This adds a constraint because age cannot decrease. 
This is captured by the transition function. 
In~\Cref{fig:example2} we see that the edges which act on feature \var{a} have now become unidirectional implying that the value of feature \var{a} cannot decrease, action \var{a-1} is not allowed. 

Defined formally, here are the components for this \textsf{MDP}:
\begin{itemize}[itemsep=0pt,topsep=2pt,leftmargin=*]
    \item States $\mathcal{S}$ = \{0,0\}, \{0,1\}, \{0,2\}, \{1,0\}, \dots. 
    \item Actions $\mathcal{A}$ = \var{a+1}, \var{b+1}, \var{b-1}. 
    \item Transition function $\mathit{T : \mathcal{S} \times \mathcal{A} \rightarrow \mathcal{S}}$ 
    \item Reward function $R : \mathcal{S} \times \mathcal{A} \to \mathbb{R}$. 
    \item Discount factor $\gamma \in [0,1)$. 
\end{itemize}

\smallskip

\begin{figure}
    \centering
    \includegraphics[width=0.8\columnwidth]{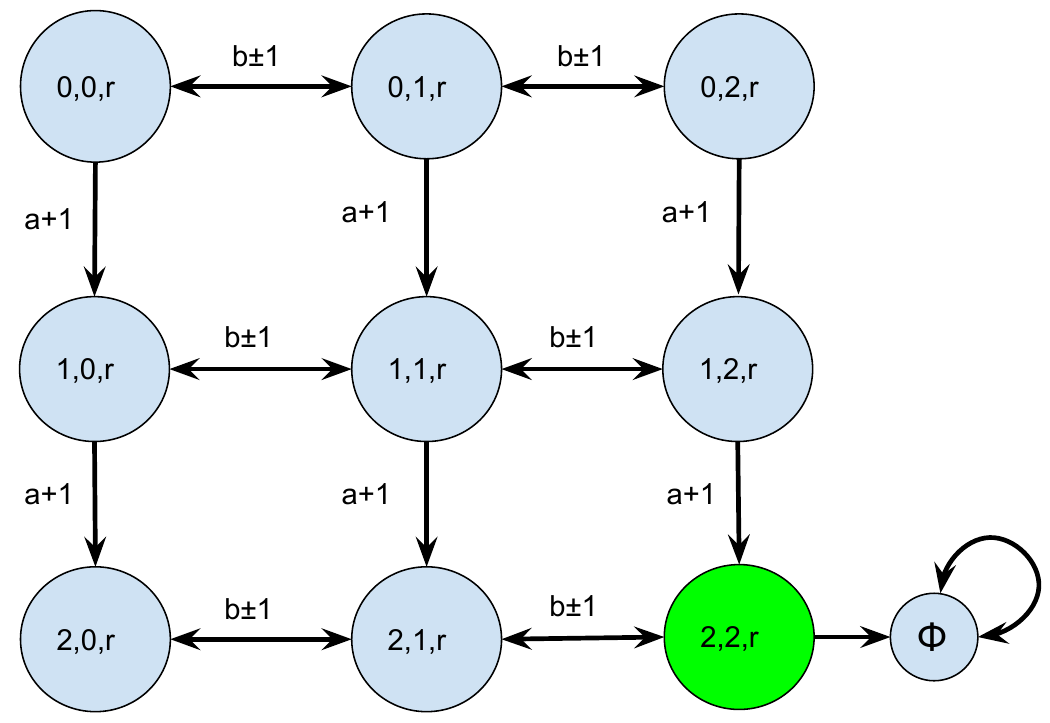}
    \caption{Transition function for a dataset with 3 features, the first being age (denoted by \var{a}) and the third being race (denoted by \var{r}). Circles show all the states and edges show possible transitions. None of the actions can change the value of feature \var{r} as race is immutable.}
    \label{fig:example3}
    \vspace{1em}
\end{figure}

\textbf{Example 2:}
Let us now consider a dataset with three features, out of which one is immutable, e.g., race (denoted by feature \var{r}). 
Feature \var{a} still represents age and carries its non-decreasing constraint. 
Such a feature cannot be changed using any action, and this is encoded in the transition function by returning the same state if this action is taken. 
The state space in this \textsf{MDP} will consist of 3 values, one for each feature. 
\Cref{fig:example3} shows the transition function for the \textsf{MDP} representing the CFE problem using this dataset. 
As we already saw, \var{a} which represents \var{age} is non-decreasing. 
Also, none of the actions affect the value of feature \var{r}, it remains constant (shown by the constant `r' in the diagram). 
The reward function is similar to the first example: a constant cost to take any action and a high reward for reaching the terminal state where the first two features are (2,2). This state follows into a dummy state where any action ends up in the same dummy state. 

Let \var{r} take values 0 and 1. 

Defined formally, here are the components for this \textsf{MDP}:
\begin{itemize}[itemsep=0pt,topsep=2pt,leftmargin=*]
    \item States $\mathcal{S}$ = \{0,0,0\}, \{0,1,0\}, \{0,2,1\}, \{1,0,0\}, \dots. 
    \item Actions $\mathcal{A}$ = \var{a+1}, \var{b+1}, \var{b-1}. 
    \item Transition function $\mathit{T : \mathcal{S} \times \mathcal{A} \rightarrow \mathcal{S}}$ 
    \item Reward function $R : \mathcal{S} \times \mathcal{A} \to \mathbb{R}$. 
    \item Discount factor $\gamma \in [0,1)$. 
\end{itemize}

\smallskip

\textbf{Example 3:}
In all the examples we visited, there was a constant cost to taking any action, and all states but one gave a 0 reward on reaching them. 
Consider the previous example where the dataset consisted of 3 features: age, education-level, and race. 
Some of the states do not appear in the training dataset used to train the classifier we are trying to generate CFEs for. 
Ideally, we would prefer to generate CFEs that are similar to existing data; otherwise, we might generate unrealistic and unactionable explanations. 
This is based on the assumption that training data is a good representation of the true distribution of features. 
Some of such states are:
\begin{itemize}[itemsep=0pt,topsep=2pt,leftmargin=*]
    \item (0,2,0) and (0,2,1): intuitively this shows that it is unrealistic for an individual to be in the lowest age group (0) and have the highest education-level (1). This is true regardless of the person's race. 
    \item (2,0,1): it is improbable for someone belonging to the race encoded by value 1 to be in the highest age group and have the lowest possible education-level. Yet (2,0,0) is not an improbable state, and this might be due to the differences in education level across different races. 
\end{itemize}
We encode this information in the \textsf{MDP} by modifying its reward function. 
If we take an action that ends up in an unrealistic state, it attracts a penalty of -5 points. 
The dummy state still carries the +10 reward, other states reward 0, and there is a constant cost of 1 to take any action. 
The agent learning in this environment would ideally learn to avoid the unrealistic states and take actions that go to the terminal state. 
In this situation, the agent can learn not to take a shorter path because it goes through an unrealistic state. 
We use a $k$-Nearest Neighbour algorithm to find the appropriate penalty for landing in any state in our experiments. If a state is close to a datapoint in the training dataset or occurs in the training dataset itself, there is a low or no penalty. 

\smallskip

\textbf{Example 4:}
Reconsider the last example in which there are three features. 
The reward function in the last example costed the same for all features. 
It might be harder to change one feature than another in real life, e.g., it might be easier for someone to wait to increase their age rather than get a higher educational level. 
This can be accounted for by posting higher costs to change features harder to change and vice-versa for feature easier to change. 

%% file: counter_vs_contrastive.tex
\section{Counterfactual vs. Contrastive explanations}
\label{sec:counter_vs_contra}

There is ongoing discussion on the exact definition of counterfactual explanation, with some researchers advocating to call it contrastive explanations. \citet{cfe_vs_contra} have captured the precise difference in a recent article. 
They mention that the counterfactual explanations as introduced by~\citet{wachter_counterfactual_2017} are almost the same as contrastive explanations. These explanations seek to find the minimal changes to the input such that the prediction from the ML model changes. 
On the other hand, counterfactuals are a function of the datapoint, its prediction, the ML model, and the data generating process that created that datapoint. 
\citet{causality:Pearl} describes three steps for generating counterfactuals: 
\begin{enumerate}
    \item Abduction: This is the process of conditioning on the exogenous variables in the data generation process. 
    \item Intervention: This is the process of making a sparse change on a specific observable variable. 
    \item Prediction: This is the process of using the exogenous variables identified in the first step and propagating the intervention to generate the counterfactual. 
\end{enumerate}
We agree with this framing.  Therefore, counterfactual explanations amount to much more perturbing the input datapoint---as in the case of contrastive explanations, which are tied to the data generating process.  Indeed, it is our belief that our proposed framework captures these concerns, if data regarding causal interactions is available.

We take note of this distinction and therefore have adherence to causal relations as a desiderata of counterfactual explanations (\Cref{sec:desiderata}). 
Structural Causal Models (SCM) consists of the exogenous and endogenous variables involved in the data generation process. 
\toolname takes as input the SCM (partial SCM is supported) of the dataset and takes it into consideration while generating CFEs.
If the SCM is not provided, the explanations generated by \toolname are basically contrastive explanations. 

%% file: algorithm_justification.tex
\section{Justification of the Choice for our Implementation Algorithm. }
\label{sec:justification}


In this section, using a set of questions and answers we attempt to justify our choice of the algorithm used by \toolname to generate CFEs. 

\Ques{Why not use planning algorithms?}
\Justify{\normalfont{Classical planning approaches generate optimal plans of reaching from the current state to the desired goal state. However, they suffer from several disadvantages compared to learning algorithms, which we enlist below:
\begin{enumerate}
    \item Most planning algorithms deal with a finite and discrete state and action space. 
    \item Most planning algorithms, e.g., random shooting, BFS, Dijkstra's, A*, Greedy, etc., need to be run for each start state (datapoint in this case) separately and hence are not amortized~\cite{planning-book}. Our baselines, Random and Greedy are example of two planning algorithms. Some planning algorithms like policy iteration are amortized though, and we discuss that in~\Qref{q:policy-iteration}. 
    \item Most planning algorithms require a goal state to be specified in order to find an optimal path to reach it. In our case, we do not know the goal state in advance. 
\end{enumerate}
}}

\Ques{Why not use Monte Carlo Tree Search (MCTS)?}
\Justify{\normalfont{MCTS takes actions at a state until a leaf node is reached, and then uses heuristics to expand among a set of leaf nodes. A value is assigned to each leaf node using a default policy and this is back-propagated to earlier nodes in the path. This way one can find a path that maximizes the reward. Similar to other planning algorithms, MCTS needs to be run separately for each datapoint and is therefore not amortized. }}

\Ques{Why not use Policy Iteration (PI)?}
\label{q:policy-iteration}
\Justify{\normalfont{Policy Iteration is an iterative algorithm to find the optimal actions to be taken at each state. As with other planning algorithms, PI can only work with a finite and discrete state and action space. Nevertheless, we used PI to learn the optimal policy for the German Credit dataset. Each numerical attribute was discretized into 4 values. Since there are 20 features in this dataset, even after sparse discretization, it resulted in a total of 141557760000 states. When the PI algorithm was run with the full state space, it requested more than 500 GB of memory. When two features were dropped, and the algorithm was run with the remaining 18 features, it still requested over 500 GB of memory. Therefore, we performed an experiment starting with 4 features and the summarized the results in~\Cref{tab:policy-iteration}. As shown in the table, the time taken to learn the optimal policy even with the subset of only 9 features is over 24 hours. We killed the processes with 15 features and higher due to extreme memory requirement, and the process with 11 features timed-out after 25 hours. Thus, PI is not able to scale for real-world datasets, both in terms of time and memory requirement. 
For a smaller subset of 4, 6, and 7 features, the learned optimal policy has 100\% validity in generating CFEs for the datapoints classified in the negative class by the ML classifier. We used the same classfier as used in the main experiments (see ~\cref{sec:eval}). 

\begin{table}[]
    \centering
    \caption{The time taken and memory consumption for the Policy Iteration algorithm with increasing number of features of the German Credit dataset. The original dataset has 20 features, and when all the features are discretized it results in a large number of states (last row in the table). Even when rerun with a subset of 9 features, the time taken to train learn the optimal policy is prohibitively long. This show that PI is not scalable for real-world datasets. }
    \label{tab:policy-iteration}
    \resizebox{\columnwidth}{!}{%
    \begin{tabular}{ccccc}
    \toprule
    \# Features & \# States  & Time Taken(s) & Memory & Validity \\
    \midrule
    4           &  320       &   67          &  0.6GB      &  100\%  \\
    6           &  1280      &   473         &  0.7GB      &  100\%  \\
    7           &  5120      &   2336        &  0.7GB      &  100\%  \\
    9           &  102400     &  86566       &  1.0GB      &  100\%  \\
    11          &  1638400      &  --        &  7.5GB      &  --  \\
    15          &  393216000      &    --     &  315GB     &  --  \\
    18          &  35389440000      &   --     &  500+GB    &  --  \\
    20 (original)   &  141557760000   &   --    & 500+GB      &  --  \\
    \bottomrule
    \end{tabular}
    }
\end{table}

}}

\Ques{Why not use Model Predictive Control (MPC)?}
\Justify{\normalfont{MPC simulates taking multiple paths upto a fixed horizon. 
It choses the path with the minimum cost but takes only one step in that path and 
repeats the simulation process to take the next step. Since it performs the path unrolling 
at each stage, it is an expensive algorithm and not amortized.}  
}

\Ques{Why not use LQR?}
\Justify{\normalfont{LQR, similar to other trajectory optimization methods, is not amortized. 
Moreover, we do not know if the cost that is coming from the black-box ML model is a quadratic function or not, and hence LQR is not applicable. }}

\Ques{Why not use model-based RL?}
\Justify{\normalfont{We don't need to learn the model for \toolname, we define it. 
Most model-based RL methods involve MPC in the planning step and therefore is not amortized. 
Model-based acceleration for model-free RL could have been used for \toolname's implementation. 
However, we choose model-free approaches for \toolname owing to availability of several 
standardized implementations on Github and other libraries~\citep{kostrikovrl,ray}.}
We leave model-based acceleration of our approach for future work. 
}

\Ques{Why not use Q-learning?}
\Justify{\normalfont{Q-learning is applicable for discrete action and state space while Policy Gradients are also applicable in continuous state and action domain. Talking about Deep Q-learning, Policy Gradients methods have been shown to have better convergence than it.}}

\Ques{Why not use vanilla Policy-Gradients?}
\Justify{\normalfont{Vanilla Policy Gradient algorithm like REINFORCE suffer from high 
variance during the Monte Carlo update. Several approaches have been proposed to tackle 
this issue and we use one of such popular approaches PPO+GAE. }}


%% file: implementation_details.tex
\section{Implementation Details of \toolname}
\label{sec:implementation-details}

\normalfont{
In this section, we expound on some of the details of how the PPO+GAE algorithm has been implemented in \toolname. 
We used the open-source implementation~\cite{kostrikovrl} for the base implementation of the PPO algorithm. 
This implementation requires the Reinforcement Learning (RL) environment to be available using the OpenAI Gym library~\cite{gym}. 
We, therefore, created the Gym environment for the datasets used in the experiments. 
We use the default approximator used in the implementation for training an RL agent using the PPO algorithm. 
Both the actor and critic are approximated by a fully connected neural network with two hidden layers; each hidden layer has 64 neurons. 
There are several hyperparameters in the PPO algorithm that can be tuned. 
We ran a moderate-size hyperparameter exploration for each dataset and selected the best one based on validity. 
\Cref{tab:finalresults} shows the results of the best hyperparameters for each dataset. 
The one-time training cost for the three datasets using this implementation was about
2 hours for the German Credit, and about 1 hour each for the Adult Income and Credit default datasets.  

We also implemented \toolname using the recently released RLib library~\cite{rlib}, 
which is build on top of the Ray distributed framework~\cite{ray}. 
Due to the distributed nature of this implementation, the training time of the 
RL agent in this framework took about 15-20 minutes for all datasets on the same CPU machine. 

For training the PPO algorithm, we experimented with different starting points for the episodes. 
In the first case, we used a random starting point sampled from the state space of the MDP. 
Note that this starting point might not be in the data manifold as it is sampled from the whole state space. 
In the second case, we used random training datapoints from the dataset as the starting point. 
It turned out that starting from the training datapoints led to better learning by the 
RL agents, and we stick to this in the implementation. 
We hypothesize that the RL agent is able to learn better actions when it starts from the training data manifold because of the loss that an agent incurs when it is far from the data manifold. 
So, for instance, even if an agent took a correct action at a starting point which is 
far off from the manifold, the next state is probably still going to be far from the manifold 
incurring a large negative reward. 
Thus when starting from off-manifold datapoints, the agent might most get negative 
rewards for all actions, and this might lead to it not learning much. 

\subsection{Consideration for fair comparison with baselines. }
The greedy baselines requires to have a finite number of actions to evaluate and greedily 
choose the best action. Therefore, the action space should be discretized. 
Therefore for a fair comparsion, we compare \toolname, random, and the greedy 
approaches using a discrete action space. 

DiCE-VAE requires several hyperparameters during training the variantional auto-encoder (VAE), which is used to generate the ARs. 
We ran a hyperparameter exploration for all the hyperparameters and reported the 
results from the best hyperparameter per dataset for DiCE-VAE. 

}

%% file: hyperparam.tex
\section{Effects of different hyperparameters on evaluation metrics}
\label{sec:hyperparam}

\begin{table*}[]
    \centering
     \caption{Comparison of the evaluation metrics of ARs for different value of 
    $\lambda$ hyper-parameter which determines the closeness to the training data manifold. }
    \resizebox{0.8\textwidth}{!}{
    \begin{tabular}{l l l l l l l l l l}
        \toprule
        Dataset   & $\lambda$ & \#DataPts. & Validity & Prox-Num & Prox-Cat & Sparsity & Manifold dist. & Causality & Time (s) \\
        \midrule
        \multirow{5}{*}{German Credit}
            &   0       & 257 & 94.6      & 0.09     & 0.060     & 1.17  & 0.70       & 100.0  & 0.12    \\
            &   0.1     & 257 & 97.3      & 0.10     & 0.063     & 1.22  & 0.72       & 100.0  & 0.07    \\
            &   1       & 257 & 95.3      & 0.11     & 0.059     & 1.20  & 0.71       & 100.0  & 0.07    \\             
            &   10      & 257 & 40.5      & 0.0      & 0.077     & 1.00  & 0.62       & 100.0  & 0.15    \\            
            &   100     & 257 & 42.4      & 0.0      & 0.079     & 1.03  & 0.64       & 100.0  & 0.22    \\           
        \midrule
        \multirow{5}{*}{Adult Income}
            &   0       & 7229 & 100.0    & 0.04     & 0.0       & 1.00  & 0.18       & 100.0  & 0.028   \\
            &   0.1     & 7229 & 100.0    & 0.04     & 0.0       & 1.00  & 0.18       & 100.0  & 0.016   \\        
            &   1       & 7229 & 100.0    & 0.04     & 0.0       & 1.00  & 0.18       & 100.0  & 0.016   \\        
            &   10      & 7229 & 100.0    & 0.04     & 0.0       & 1.00  & 0.18       & 100.0  & 0.016   \\
            &   100     & 7229 & 100.0    & 0.04     & 0.0       & 1.00  & 0.18       & 100.0  & 0.028   \\
        \midrule
        \multirow{5}{*}{Credit Default}
            &   0        & 5363 & 99.96    & 0.007    & 0.11     & 1.00  & 0.32      & 100.0  & 0.08    \\
            &   0.1      & 5363 & 99.96    & 0.015    & 0.11     & 1.00  & 0.32      & 100.0  & 0.05   \\
            &   1        & 5363 & 79.38    & 0.001    & 0.14     & 1.28  & 0.37      & 100.0  & 0.08   \\
            &   10       & 5363 & 47.29    & 0.44     & 0.0      & 1.00  & 0.24      & 100.0  & 0.18   \\
            &   100      & 5363 & 5.74     & 1.22     & 0.0      & 1.00  & 0.60      & 100.0  & 0.32   \\
            
        \bottomrule
    \end{tabular}
    }
   
    \label{tab:lambda_effects}
\end{table*}

\normalfont{
\Cref{tab:lambda_effects} shows the effect of increasing the penalty for 
leaving the training data manifold, which is enforced at each step of a counterfactual path. 
With increasing $\lambda$, the manifold distance should become smaller.
Increasing $\lambda$ also makes it harder for the agent to learn a effective policy, 
and therefore the validity could also go down. 
We observe both these expected trends in \Cref{tab:lambda_effects} above, 
specially for the German Credit and Credit Default datasets. 
}

%% file: aaai22.bbl
\begin{thebibliography}{74}
\providecommand{\natexlab}[1]{#1}

\bibitem[{Adadi and Berrada(2018)}]{xai-survey2}
Adadi, A.; and Berrada, M. 2018.
\newblock Peeking inside the black-box: A survey on Explainable Artificial
  Intelligence (XAI).
\newblock \emph{IEEE Access}, PP.

\bibitem[{Andrews, Diederich, and Tickle(1995)}]{Andrews_exp6}
Andrews, R.; Diederich, J.; and Tickle, A.~B. 1995.
\newblock Survey and Critique of Techniques for Extracting Rules from Trained
  Artificial Neural Networks.
\newblock \emph{Know.-Based Syst.}, 8.

\bibitem[{Brockman et~al.(2016)Brockman, Cheung, Pettersson, Schneider,
  Schulman, Tang, and Zaremba}]{gym}
Brockman, G.; Cheung, V.; Pettersson, L.; Schneider, J.; Schulman, J.; Tang,
  J.; and Zaremba, W. 2016.
\newblock OpenAI Gym.
\newblock arXiv:arXiv:1606.01540.

\bibitem[{Carvalho, Pereira, and Cardoso(2019)}]{carvalho2019:survey3}
Carvalho, D.~V.; Pereira, E.~M.; and Cardoso, J.~S. 2019.
\newblock Machine learning interpretability: A survey on methods and metrics.
\newblock \emph{Electronics}, 8.

\bibitem[{Chipman, George, and Mcculloch(1998)}]{Chipman_makingsense_exp3}
Chipman, H.~A.; George, E.~I.; and Mcculloch, R.~E. 1998.
\newblock Making Sense of a Forest of Trees.
\newblock In \emph{Proceedings of the 30th Symposium on the Interface}.

\bibitem[{Chou et~al.(2021)Chou, Moreira, Bruza, Ouyang, and
  Jorge}]{Chou21:Counterfactuals}
Chou, Y.-L.; Moreira, C.; Bruza, P.; Ouyang, C.; and Jorge, J. 2021.
\newblock Counterfactuals and Causability in Explainable Artificial
  Intelligence: Theory, Algorithms, and Applications.
\newblock \emph{arXiv preprint arXiv:2103.04244}.

\bibitem[{Craven and Shavlik(1995)}]{craven_exp1}
Craven, M.~W.; and Shavlik, J.~W. 1995.
\newblock Extracting Tree-Structured Representations of Trained Networks.
\newblock In \emph{Conference on Neural Information Processing Systems
  (NeurIPS)}. Cambridge, MA, USA: MIT Press.

\bibitem[{Dandl et~al.(2020)Dandl, Molnar, Binder, and
  Bischl}]{dandl_multi-objective_2020}
Dandl, S.; Molnar, C.; Binder, M.; and Bischl, B. 2020.
\newblock Multi-{Objective} {Counterfactual} {Explanations}.
\newblock \emph{arXiv:2004.11165 [cs, stat]}.

\bibitem[{Deng(2014)}]{Deng_exp5}
Deng, H. 2014.
\newblock Interpreting Tree Ensembles with inTrees.
\newblock \emph{arXiv:1408.5456}.

\bibitem[{Dhurandhar et~al.(2018)Dhurandhar, Chen, Luss, Tu, Ting, Shanmugam,
  and Das}]{dhurandhar_explanations_2018}
Dhurandhar, A.; Chen, P.-Y.; Luss, R.; Tu, C.-C.; Ting, P.; Shanmugam, K.; and
  Das, P. 2018.
\newblock Explanations Based on the Missing: Towards Contrastive Explanations
  with Pertinent Negatives.
\newblock In \emph{Conference on Neural Information Processing Systems
  (NeurIPS)}.

\bibitem[{Dhurandhar et~al.(2019)Dhurandhar, Pedapati, Balakrishnan, Chen,
  Shanmugam, and Puri}]{dhurandhar_model_2019}
Dhurandhar, A.; Pedapati, T.; Balakrishnan, A.; Chen, P.-Y.; Shanmugam, K.; and
  Puri, R. 2019.
\newblock Model {Agnostic} {Contrastive} {Explanations} for {Structured}
  {Data}.
\newblock \emph{arXiv:1906.00117 [cs, stat]}.

\bibitem[{Dhurandhar and Shanmugam(2020)}]{cfe_vs_contra}
Dhurandhar, A.; and Shanmugam, K. 2020.
\newblock Counterfactual vs Contrastive Explanations in Artificial
  Intelligence.
\newblock
  \url{https://towardsdatascience.com/counterfactual-vs-contrastive-explanations-in-artificial-intelligence-e67a9cfc7e4e}.
\newblock Accessed: 2021-05-15.

\bibitem[{Domingos(1998)}]{Pedro_exp4}
Domingos, P. 1998.
\newblock Knowledge Discovery Via Multiple Models.
\newblock \emph{Intell. Data Anal.}, 2.

\bibitem[{Dua and Graff(2017)}]{UCI-repo}
Dua, D.; and Graff, C. 2017.
\newblock {UCI} Machine Learning Repository.

\bibitem[{Dunkelau and Leuschel(2019)}]{dunkelau_fairness-aware}
Dunkelau, J.; and Leuschel, M. 2019.
\newblock Fairness-{Aware} {Machine} {Learning}.

\bibitem[{Ehsan et~al.(2021{\natexlab{a}})Ehsan, Liao, Muller, Riedl, and
  Weisz}]{ehsan2021expanding}
Ehsan, U.; Liao, Q.~V.; Muller, M.; Riedl, M.~O.; and Weisz, J.~D.
  2021{\natexlab{a}}.
\newblock Expanding Explainability: Towards Social Transparency in AI systems.
\newblock In \emph{CHI}.

\bibitem[{Ehsan et~al.(2021{\natexlab{b}})Ehsan, Wintersberger, Liao, Mara,
  Streit, Wachter, Riener, and Riedl}]{Ehsan21:Operationalizing}
Ehsan, U.; Wintersberger, P.; Liao, Q.~V.; Mara, M.; Streit, M.; Wachter, S.;
  Riener, A.; and Riedl, M.~O. 2021{\natexlab{b}}.
\newblock Operationalizing Human-Centered Perspectives in Explainable AI.
\newblock In \emph{CHI}.

\bibitem[{Faggella(2020)}]{medical-treatment-ml}
Faggella, D. 2020.
\newblock Machine Learning for Medical Diagnostics – 4 Current Applications.
\newblock
  \url{https://emerj.com/ai-sector-overviews/machine-learning-medical-diagnostics-4-current-applications/}.
\newblock Accessed: 2020-10-15.

\bibitem[{Ghallab, Nau, and Traverso(2016)}]{planning-book}
Ghallab, M.; Nau, D.; and Traverso, P. 2016.
\newblock \emph{Automated Planning and Acting}.
\newblock USA: Cambridge University Press, 1st edition.
\newblock ISBN 1107037271.

\bibitem[{Grath et~al.(2018)Grath, Costabello, Van, Sweeney, Kamiab, Shen, and
  Lecue}]{grath_interpretable_2018}
Grath, R.~M.; Costabello, L.; Van, C.~L.; Sweeney, P.; Kamiab, F.; Shen, Z.;
  and Lecue, F. 2018.
\newblock Interpretable {Credit} {Application} {Predictions} {With}
  {Counterfactual} {Explanations}.
\newblock \emph{arXiv:1811.05245 [cs]}.

\bibitem[{Guidotti et~al.(2018{\natexlab{a}})Guidotti, Monreale, Ruggieri,
  Pedreschi, Turini, and Giannotti}]{guidotti_local_2018}
Guidotti, R.; Monreale, A.; Ruggieri, S.; Pedreschi, D.; Turini, F.; and
  Giannotti, F. 2018{\natexlab{a}}.
\newblock Local {Rule}-{Based} {Explanations} of {Black} {Box} {Decision}
  {Systems}.

\bibitem[{Guidotti et~al.(2018{\natexlab{b}})Guidotti, Monreale, Ruggieri,
  Turini, Giannotti, and Pedreschi}]{xai-survey4}
Guidotti, R.; Monreale, A.; Ruggieri, S.; Turini, F.; Giannotti, F.; and
  Pedreschi, D. 2018{\natexlab{b}}.
\newblock A Survey of Methods for Explaining Black Box Models.
\newblock \emph{ACM Comput. Surv.}, 51.

\bibitem[{Henelius et~al.(2014)Henelius, Puolam\"{a}ki, Bostr\"{o}m, Asker, and
  Papapetrou}]{Andreas_exp7}
Henelius, A.; Puolam\"{a}ki, K.; Bostr\"{o}m, H.; Asker, L.; and Papapetrou, P.
  2014.
\newblock A Peek into the Black Box: Exploring Classifiers by Randomization.
\newblock \emph{Data Min. Knowl. Discov.}, 28.

\bibitem[{Holstein et~al.(2019)Holstein, Wortman~Vaughan, Daum{\'e}~III, Dudik,
  and Wallach}]{Holstein19:Improving}
Holstein, K.; Wortman~Vaughan, J.; Daum{\'e}~III, H.; Dudik, M.; and Wallach,
  H. 2019.
\newblock Improving fairness in machine learning systems: What do industry
  practitioners need?
\newblock In \emph{Conference on Human Factors in Computing Systems (CHI)},
  1--16.

\bibitem[{Joshi et~al.(2019)Joshi, Koyejo, Vijitbenjaronk, Kim, and
  Ghosh}]{joshi_towards_2019}
Joshi, S.; Koyejo, O.; Vijitbenjaronk, W.; Kim, B.; and Ghosh, J. 2019.
\newblock Towards {Realistic} {Individual} {Recourse} and {Actionable}
  {Explanations} in {Black}-{Box} {Decision} {Making} {Systems}.

\bibitem[{Kanamori et~al.(2020)Kanamori, Takagi, Kobayashi, and
  Arimura}]{Kanamori2020:DACE}
Kanamori, K.; Takagi, T.; Kobayashi, K.; and Arimura, H. 2020.
\newblock DACE: Distribution-Aware Counterfactual Explanation by Mixed-Integer
  Linear Optimization.
\newblock In \emph{International Joint Conference on Artificial Intelligence
  (IJCAI)}.

\bibitem[{Karimi et~al.(2020{\natexlab{a}})Karimi, Barthe, Balle, and
  Valera}]{karimi_model-agnostic_2020}
Karimi, A.-H.; Barthe, G.; Balle, B.; and Valera, I. 2020{\natexlab{a}}.
\newblock Model-Agnostic Counterfactual Explanations for Consequential
  Decisions.
\newblock In \emph{Proceedings of the 23rd International Conference on
  Artificial Intelligence and Statistics}.

\bibitem[{Karimi et~al.(2020{\natexlab{b}})Karimi, Barthe, Schölkopf, and
  Valera}]{karimi2020survey}
Karimi, A.-H.; Barthe, G.; Schölkopf, B.; and Valera, I. 2020{\natexlab{b}}.
\newblock A survey of algorithmic recourse: definitions, formulations,
  solutions, and prospects.
\newblock arXiv:2010.04050.

\bibitem[{Karimi, Schölkopf, and Valera(2020)}]{karimi_algorithmic_2020}
Karimi, A.-H.; Schölkopf, B.; and Valera, I. 2020.
\newblock Algorithmic {Recourse}: from {Counterfactual} {Explanations} to
  {Interventions}.
\newblock \emph{arXiv:2002.06278 [cs, stat]}.

\bibitem[{Karimi et~al.(2020{\natexlab{c}})Karimi, von Kügelgen, Schölkopf,
  and Valera}]{karimi-imperfect:2020}
Karimi, A.-H.; von Kügelgen, J.; Schölkopf, B.; and Valera, I.
  2020{\natexlab{c}}.
\newblock Algorithmic recourse under imperfect causal knowledge: a
  probabilistic approach.

\bibitem[{Keane and Smyth(2020)}]{keane2020good}
Keane, M.~T.; and Smyth, B. 2020.
\newblock Good Counterfactuals and Where to Find Them: A Case-Based Technique
  for Generating Counterfactuals for Explainable AI (XAI).
\newblock arXiv:2005.13997.

\bibitem[{Kostrikov(2018)}]{kostrikovrl}
Kostrikov, I. 2018.
\newblock PyTorch Implementations of Reinforcement Learning Algorithms.
\newblock \url{https://github.com/ikostrikov/pytorch-a2c-ppo-acktr-gail}.

\bibitem[{Krishnan, Sivakumar, and Bhattacharya(1999)}]{KRISHNAN_exp2}
Krishnan, R.; Sivakumar, G.; and Bhattacharya, P. 1999.
\newblock Extracting decision trees from trained neural networks.
\newblock \emph{Pattern Recognition}, 32.

\bibitem[{Krishnan and Wu(2017)}]{Krishnan_exp8}
Krishnan, S.; and Wu, E. 2017.
\newblock PALM: Machine Learning Explanations For Iterative Debugging.
\newblock In \emph{Proceedings of the 2nd Workshop on Human-In-the-Loop Data
  Analytics}.

\bibitem[{Lash et~al.(2017)Lash, Lin, Street, Robinson, and
  Ohlmann}]{inverse-classification2}
Lash, M.~T.; Lin, Q.; Street, W.~N.; Robinson, J.~G.; and Ohlmann, J.~W. 2017.
\newblock Generalized Inverse Classification.
\newblock In \emph{SDM}.

\bibitem[{Laugel et~al.(2018)Laugel, Lesot, Marsala, Renard, and
  Detyniecki}]{medina_comparison-based_2018}
Laugel, T.; Lesot, M.-J.; Marsala, C.; Renard, X.; and Detyniecki, M. 2018.
\newblock Comparison-{Based} {Inverse} {Classification} for {Interpretability}
  in {Machine} {Learning}.
\newblock In \emph{Information Processing and Management of Uncertainty in
  Knowledge-Based Systems, Theory and Foundations (IPMU)}. Springer
  International Publishing.

\bibitem[{Le, Wang, and Lee(2019)}]{Grace:2019}
Le, T.; Wang, S.; and Lee, D. 2019.
\newblock GRACE: Generating Concise and Informative Contrastive Sample to
  Explain Neural Network Model's Prediction.
\newblock arXiv:1911.02042.

\bibitem[{Liang et~al.(2018)Liang, Liaw, Nishihara, Moritz, Fox, Goldberg,
  Gonzalez, Jordan, and Stoica}]{rlib}
Liang, E.; Liaw, R.; Nishihara, R.; Moritz, P.; Fox, R.; Goldberg, K.;
  Gonzalez, J.~E.; Jordan, M.~I.; and Stoica, I. 2018.
\newblock RLlib: Abstractions for Distributed Reinforcement Learning.
\newblock In \emph{ICML}.

\bibitem[{Liang et~al.(2020)Liang, Nishihara, Liaw, and et~al.}]{ray}
Liang, E.; Nishihara, R.; Liaw, R.; and et~al. 2020.
\newblock Ray.
\newblock \url{https://github.com/ray-project/ray}.

\bibitem[{Mahajan, Tan, and Sharma(2020)}]{mahajan_preserving_2020}
Mahajan, D.; Tan, C.; and Sharma, A. 2020.
\newblock Preserving {Causal} {Constraints} in {Counterfactual} {Explanations}
  for {Machine} {Learning} {Classifiers}.
\newblock \emph{arXiv:1912.03277 [cs, stat]}.

\bibitem[{Miller(2019)}]{Miller-xai:2019}
Miller, T. 2019.
\newblock Explanation in artificial intelligence: Insights from the social
  sciences.
\newblock \emph{Artificial Intelligence}, 267.

\bibitem[{Mnih et~al.(2016)Mnih, Badia, Mirza, Graves, Lillicrap, Harley,
  Silver, and Kavukcuoglu}]{a3c-mniha16}
Mnih, V.; Badia, A.~P.; Mirza, M.; Graves, A.; Lillicrap, T.; Harley, T.;
  Silver, D.; and Kavukcuoglu, K. 2016.
\newblock Asynchronous Methods for Deep Reinforcement Learning.
\newblock In \emph{Proceedings of The 33rd International Conference on Machine
  Learning}. PMLR.

\bibitem[{Mothilal, Sharma, and Tan(2020)}]{mothilal_explaining_2020}
Mothilal, R.~K.; Sharma, A.; and Tan, C. 2020.
\newblock Explaining Machine Learning Classifiers through Diverse
  Counterfactual Explanations.
\newblock In \emph{Proceedings of the Conference on Fairness, Accountability,
  and Transparency (FAccT)}, FAT* ’20.

\bibitem[{Naumann and Ntoutsi(2021)}]{naumann2021sequential}
Naumann, P.; and Ntoutsi, E. 2021.
\newblock Consequence-aware Sequential Counterfactual Generation.
\newblock arXiv:2104.05592.

\bibitem[{Pawelczyk, Broelemann, and Kasneci(2020)}]{pawelczyk_learning_2020}
Pawelczyk, M.; Broelemann, K.; and Kasneci, G. 2020.
\newblock Learning {Model}-{Agnostic} {Counterfactual} {Explanations} for
  {Tabular} {Data}.
\newblock \emph{Proceedings of The Web Conference (WWW)}.

\bibitem[{Pearl(2000)}]{causality:Pearl}
Pearl, J. 2000.
\newblock \emph{Causality: Models, Reasoning, and Inference}.
\newblock USA: Cambridge University Press.
\newblock ISBN 0521773628.

\bibitem[{Poulin et~al.(2006)Poulin, Eisner, Szafron, Lu, Greiner, Wishart,
  Fyshe, Pearcy, MacDonell, and Anvik}]{Poulin_explaind}
Poulin, B.; Eisner, R.; Szafron, D.; Lu, P.; Greiner, R.; Wishart, D.~S.;
  Fyshe, A.; Pearcy, B.; MacDonell, C.; and Anvik, J. 2006.
\newblock Visual Explanation of Evidence in Additive Classifiers.
\newblock In \emph{Conference on Innovative Applications of Artificial
  Intelligence (IAAI)}.

\bibitem[{Poursabzi et~al.(2021)Poursabzi, Goldstein, Hofman, Wortman~Vaughan,
  and Wallach}]{poursabzi2021manipulating}
Poursabzi, F.~S.; Goldstein, D.~G.; Hofman, J.~M.; Wortman~Vaughan, J.~W.; and
  Wallach, H. 2021.
\newblock Manipulating and measuring model interpretability.
\newblock In \emph{CHI}.

\bibitem[{Poyiadzi et~al.(2020)Poyiadzi, Sokol, Santos-Rodriguez, De~Bie, and
  Flach}]{poyiadzi_face_2020}
Poyiadzi, R.; Sokol, K.; Santos-Rodriguez, R.; De~Bie, T.; and Flach, P. 2020.
\newblock {FACE}: {Feasible} and {Actionable} {Counterfactual} {Explanations}.
\newblock \emph{Conference on Artificial Intelligence (AAAI)}.

\bibitem[{Ramakrishnan, Lee, and
  Albarghouthi(2020)}]{Ramakrishnan_Lee_Albarghouthi_2020}
Ramakrishnan, G.; Lee, Y.~C.; and Albarghouthi, A. 2020.
\newblock Synthesizing Action Sequences for Modifying Model Decisions.
\newblock \emph{Proceedings of the AAAI Conference on Artificial Intelligence},
  34.

\bibitem[{Rathi(2019)}]{rathi-generating:2019}
Rathi, S. 2019.
\newblock Generating {Counterfactual} and {Contrastive} {Explanations} using
  {SHAP}.

\bibitem[{Ribeiro, Singh, and Guestrin(2016)}]{ribeiro_why_2016}
Ribeiro, M.~T.; Singh, S.; and Guestrin, C. 2016.
\newblock "Why Should I Trust You?": Explaining the Predictions of Any
  Classifier.
\newblock In \emph{Proceedings of the 22nd ACM SIGKDD International Conference
  on Knowledge Discovery and Data Mining}, KDD '16. New York, NY, USA:
  Association for Computing Machinery.

\bibitem[{Rudin(2019)}]{Rudin19:Stop}
Rudin, C. 2019.
\newblock Stop explaining black box machine learning models for high stakes
  decisions and use interpretable models instead.
\newblock \emph{Nature Machine Intelligence}, 1(5).

\bibitem[{Russell(2019)}]{russell_efficient_2019}
Russell, C. 2019.
\newblock Efficient Search for Diverse Coherent Explanations.
\newblock In \emph{Proceedings of the Conference on Fairness, Accountability,
  and Transparency (FAccT)}, FAT* ’19.

\bibitem[{Saha et~al.(2020)Saha, Schumann, Mcelfresh, Dickerson, Mazurek, and
  Tschantz}]{Saha20:Measuring}
Saha, D.; Schumann, C.; Mcelfresh, D.; Dickerson, J.; Mazurek, M.; and
  Tschantz, M. 2020.
\newblock Measuring non-expert comprehension of machine learning fairness
  metrics.
\newblock In \emph{International Conference on Machine Learning (ICML)}.

\bibitem[{Schulman et~al.(2018)Schulman, Moritz, Levine, Jordan, and
  Abbeel}]{GAE2018}
Schulman, J.; Moritz, P.; Levine, S.; Jordan, M.; and Abbeel, P. 2018.
\newblock High-Dimensional Continuous Control Using Generalized Advantage
  Estimation.
\newblock arXiv:1506.02438.

\bibitem[{Schulman et~al.(2017)Schulman, Wolski, Dhariwal, Radford, and
  Klimov}]{PPO2017}
Schulman, J.; Wolski, F.; Dhariwal, P.; Radford, A.; and Klimov, O. 2017.
\newblock Proximal Policy Optimization Algorithms.
\newblock arXiv:1707.06347.

\bibitem[{{Selvaraju} et~al.(2017){Selvaraju}, {Cogswell}, {Das}, {Vedantam},
  {Parikh}, and {Batra}}]{grad-cam}
{Selvaraju}, R.~R.; {Cogswell}, M.; {Das}, A.; {Vedantam}, R.; {Parikh}, D.;
  and {Batra}, D. 2017.
\newblock Grad-CAM: Visual Explanations from Deep Networks via Gradient-Based
  Localization.
\newblock In \emph{IEEE International Conference on Computer Vision}.

\bibitem[{Sennaar(2019)}]{hiring-ml}
Sennaar, K. 2019.
\newblock Machine Learning for Recruiting and Hiring – 6 Current
  Applications.
\newblock
  \url{https://emerj.com/ai-sector-overviews/machine-learning-for-recruiting-and-hiring/}.
\newblock Accessed: 2020-10-15.

\bibitem[{Sharma, Henderson, and Ghosh(2019)}]{sharma_certifai_2019}
Sharma, S.; Henderson, J.; and Ghosh, J. 2019.
\newblock {CERTIFAI}: {Counterfactual} {Explanations} for {Robustness},
  {Transparency}, {Interpretability}, and {Fairness} of {Artificial}
  {Intelligence} models.

\bibitem[{Singh et~al.(2021)Singh, Dourish, Howe, Miller, Sonenberg, Velloso,
  and Vetere}]{singh_directive_2021}
Singh, R.; Dourish, P.; Howe, P.; Miller, T.; Sonenberg, L.; Velloso, E.; and
  Vetere, F. 2021.
\newblock Directive {Explanations} for {Actionable} {Explainability} in
  {Machine} {Learning} {Applications}.
\newblock arXiv:arXiv:2102.02671.

\bibitem[{Singla(2020)}]{credit-risk-ml}
Singla, S. 2020.
\newblock Machine Learning to Predict Credit Risk in Lending Industry.
\newblock
  \url{https://www.aitimejournal.com/@saurav.singla/machine-learning-to-predict-credit-risk-in-lending-industry}.
\newblock Accessed: 2020-10-15.

\bibitem[{Sutton and Barto(2018)}]{RLBook}
Sutton, R.~S.; and Barto, A.~G. 2018.
\newblock \emph{Reinforcement Learning: An Introduction}.
\newblock Cambridge, MA, USA: A Bradford Book.
\newblock ISBN 0262039249.

\bibitem[{Tashea(2017)}]{parole-ml}
Tashea, J. 2017.
\newblock Courts Are Using AI to Sentence Criminals. That Must Stop Now.
\newblock
  \url{https://www.wired.com/2017/04/courts-using-ai-sentence-criminals-must-stop-now/}.
\newblock Accessed: 2020-10-15.

\bibitem[{Tjoa and Guan(2019)}]{tjoa2019survey1}
Tjoa, E.; and Guan, C. 2019.
\newblock A Survey on Explainable Artificial Intelligence (XAI): Towards
  Medical XAI.
\newblock arXiv:1907.07374.

\bibitem[{Turner(2016)}]{Turner2016_MES}
Turner, R. 2016.
\newblock A Model Explanation System: Latest Updates and Extensions.

\bibitem[{Ustun, Spangher, and Liu(2019)}]{Ustun19:Actionable}
Ustun, B.; Spangher, A.; and Liu, Y. 2019.
\newblock Actionable recourse in linear classification.
\newblock In \emph{Proceedings of the Conference on Fairness, Accountability,
  and Transparency (FAccT)}.

\bibitem[{Van~Looveren and Klaise(2020)}]{van_looveren_interpretable_2020}
Van~Looveren, A.; and Klaise, J. 2020.
\newblock Interpretable {Counterfactual} {Explanations} {Guided} by
  {Prototypes}.

\bibitem[{Verma, Dickerson, and Hines(2020)}]{verma2020CFsurvey}
Verma, S.; Dickerson, J.; and Hines, K. 2020.
\newblock Counterfactual Explanations for Machine Learning: A Review.
\newblock arXiv:2010.10596.

\bibitem[{Verma and Rubin(2018)}]{verma_fairness}
Verma, S.; and Rubin, J. 2018.
\newblock Fairness Definitions Explained.
\newblock In \emph{Proceedings of the International Workshop on Software
  Fairness}, FairWare ’18. New York, NY, USA: Association for Computing
  Machinery.

\bibitem[{Wachter, Mittelstadt, and
  Russell(2017)}]{wachter_counterfactual_2017}
Wachter, S.; Mittelstadt, B.; and Russell, C. 2017.
\newblock Counterfactual {Explanations} {Without} {Opening} the {Black} {Box}:
  {Automated} {Decisions} and the {GDPR}.
\newblock \emph{SSRN Electronic Journal}.

\bibitem[{White and Garcez(2019)}]{white_measurable_2019}
White, A.; and Garcez, A.~d. 2019.
\newblock Measurable {Counterfactual} {Local} {Explanations} for {Any}
  {Classifier}.
\newblock \emph{arXiv:1908.03020 [cs]}.

\bibitem[{{Zhou} et~al.(2016){Zhou}, {Khosla}, {Lapedriza}, {Oliva}, and
  {Torralba}}]{Khosla_cam}
{Zhou}, B.; {Khosla}, A.; {Lapedriza}, A.; {Oliva}, A.; and {Torralba}, A.
  2016.
\newblock Learning Deep Features for Discriminative Localization.
\newblock In \emph{CVPR}.

\bibitem[{Zien et~al.(2009)Zien, Kr{\"a}mer, Sonnenburg, and
  R{\"a}tsch}]{Zien_exp9}
Zien, A.; Kr{\"a}mer, N.; Sonnenburg, S.; and R{\"a}tsch, G. 2009.
\newblock The Feature Importance Ranking Measure.
\newblock In \emph{Machine Learning and Knowledge Discovery in Databases}.
  Berlin, Heidelberg: Springer Berlin Heidelberg.

\end{thebibliography}
